\documentclass[lettersize,journal]{IEEEtran}
\usepackage{amsmath,amsfonts}
\usepackage{algorithmic}
\usepackage{algorithm}
\usepackage{array}
\usepackage[caption=false,font=normalsize,labelfont=sf,textfont=sf]{subfig}
\usepackage{textcomp}
\usepackage{stfloats}
\usepackage{url}
\usepackage{verbatim}
\usepackage{graphicx}
\usepackage{cite}
\usepackage{hyperref}
\usepackage{amssymb}
\usepackage{booktabs}
\usepackage{color}
\usepackage{multirow}
\usepackage{utfsym}

\definecolor{light_blue}{rgb}{0.867,0.922,0.969}
\definecolor{purple}{rgb}{0.7,0,0.7}
\definecolor{red}{rgb}{1.0,0,0.0}

\newcommand{\ie}{\emph{i.e.}}
\newcommand{\eg}{\emph{e.g.}}

\begin{document}

\title{Towards Unified 3D Object Detection via Algorithm and Data Unification}

\noindent \author{Zhuoling Li$^{1}$, Xiaogang Xu$^{2}$, Ser-Nam Lim$^{3}$, Hengshuang Zhao$^{1}$
\thanks{
$^{1}$Zhuoling Li and Hengshuang Zhao are with The University of Hong Kong.
$^{2}$Xiaogang Xu is with The Chinese University of Hong Kong. 
$^{3}$Ser-Nam Lim is with University of Central Florida. 
Corresponding author: Hengshuang Zhao {\tt\footnotesize (e-mail: hszhao@cs.hku.hk). Project page is at \href{https://lizhuoling.github.io/UniMODE_webpage/}{here}}.
}
}

\markboth{Journal of \LaTeX\ Class Files,~Vol.~14, No.~8, August~2021}%
{Shell \MakeLowercase{\textit{et al.}}: A Sample Article Using IEEEtran.cls for IEEE Journals}


\maketitle

\begin{abstract}
Realizing unified 3D object detection, including both indoor and outdoor scenes, holds great importance in applications like robot navigation. However, involving various scenarios of data to train models poses challenges due to their significantly distinct characteristics, \eg, diverse geometry properties and heterogeneous domain distributions. In this work, we propose to address the challenges from two perspectives, the algorithm perspective and data perspective. In terms of the algorithm perspective, we first build a monocular 3D object detector based on the bird's-eye-view (BEV) detection paradigm, where the explicit feature projection is beneficial to addressing the geometry learning ambiguity. In this detector, we split the classical BEV detection architecture into two stages and propose an uneven BEV grid design to handle the convergence instability caused by geometry difference between scenarios. Besides, we develop a sparse BEV feature projection strategy to reduce the computational cost and a unified domain alignment method to handle heterogeneous domains. From the data perspective, we propose to incorporate depth information to improve training robustness. Specifically, we build the first unified multi-modal 3D object detection benchmark MM-Omni3D and extend the aforementioned monocular detector to its multi-modal version, which is the first unified multi-modal 3D object detector. We name the designed monocular and multi-modal detectors as UniMODE and MM-UniMODE, respectively. The experimental results reveal several insightful findings highlighting the benefits of multi-modal data and confirm the effectiveness of all the proposed strategies. 
\end{abstract}

\begin{IEEEkeywords}
3D Object Detection, Unified Detection, Monocular Detection, Multi-modal Detection Benchmark.
\end{IEEEkeywords}

\section{Introduction}
\label{Sec: Introduction}

As a fundamental perception task, 3D object detection is to help agents understand the real world. Over the years, it has gained substantial interest from both the academia and industrial communities, leading to the development of numerous 3D object detectors \cite{li2024unimode,li2023bevdepth,li2022bevformer,liu2022petr,huang2021bevdet}. However, most of these detectors are trained and evaluated within a single scenario, essentially confining them to a limited number of domains. Among the detectors, many are designed for outdoor scenarios such as urban driving \cite{li2021monocular,zhang2021objects}, and the others focus on indoor detection \cite{rukhovich2022imvoxelnet}. In contrast, numerous applications such as robotic navigation, require detectors to perform well in diverse environments, including both indoor and outdoor scenarios \cite{goertzel2014artificial}. This highlights the need for exploring more unified detectors.

\begin{figure*}[tbp]
    \centering
    \includegraphics[width=\linewidth]{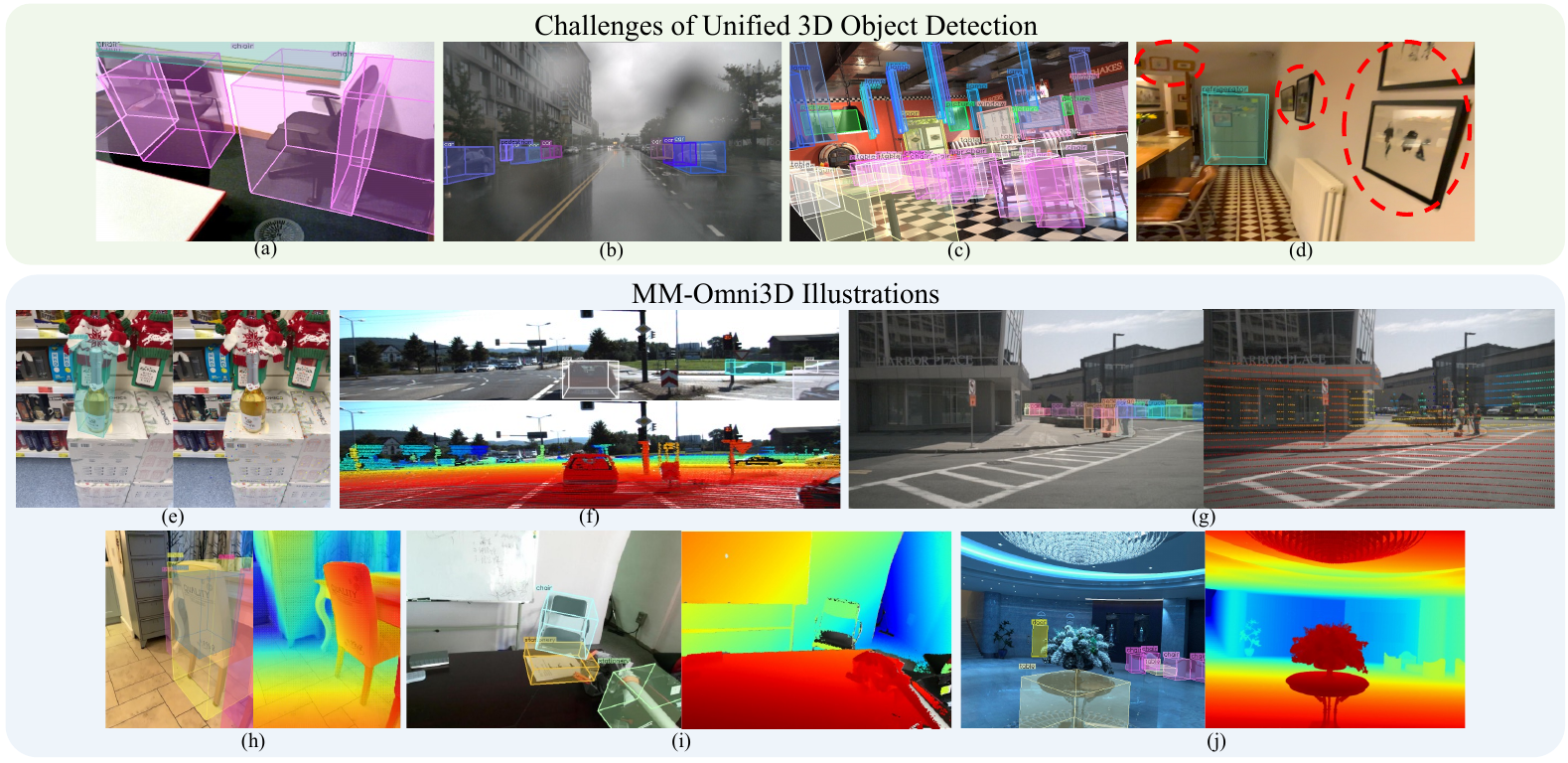}
    \caption{Sub-figures (a)$\sim$(d): Challenges of unified 3D object detection. (1) Comparing sub-figures (a) and (b), indoor objects are small and close, while outdoor objects are far and sparse. Besides, the camera parameters are highly varying. (2) Comparing sub-figures (a), (b), and (c), which correspond to a real-world indoor image, a real-world outdoor image, and a synthetic indoor image, the image styles are different. (3) Although the category ``Picture" is labeled in sub-figure (c), it is not labeled in sub-figure (d), which suggests label conflict among different sub-datasets. Unlabeled objects are highlighted by red ellipses.
    Sub-figures (e)$\sim$(j): Illustrations of the MM-Omni3D benchmark, which showcase both the 3D box annotations and point clouds. The sub-figures clearly demonstrate the significant point cloud differences between different scenarios due to depth sensor discrepancies.} \label{Fig: teaser}
    \vspace{-0.1cm}
\end{figure*}

The most critical challenge in unified 3D object detection lies in addressing the distinct characteristics of different scenarios. For example, indoor objects are smaller and closer in proximity, while outdoor detection needs to cover a vast perception range. Recently, Cube RCNN~\cite{brazil2023omni3d} has served as a predecessor in studying this problem. It directly produces 3D box predictions in the camera view and adopts a depth decoupling strategy to tackle the domain gap among scenes. However, we observe that it suffers serious convergence difficulty during training. To overcome the unstable convergence of Cube RCNN, we employ the recent popular bird's-eye-view (BEV) detection paradigm to develop a unified 3D object detector. This is because the feature projection in the BEV paradigm aligns the image space with the 3D real-world space explicitly \cite{li2022bevformer}, which alleviates the learning ambiguity in monocular 3D object detection. Nevertheless, after extensive exploration, we find that naively adopting existing BEV detection architectures~\cite{li2022bevformer,liu2022petr} does not yield promising performance, which is mainly blamed on the following obstacles.

First of all, as shown in  Fig.~\ref{Fig: teaser} (a) and (b), the geometry properties (\eg, perception ranges, target positions) between indoor and outdoor scenes are diverse. Specifically, indoor objects are typically a few meters away from the camera, while outdoor targets can be more than 100m away. Since a unified BEV detector needs to recognize objects in all scenarios, the BEV feature has to cover the maximum possible perception range. Meanwhile, as indoor objects are often small, the BEV grid resolution for indoor detection should be precise. All these characteristics lead to unstable convergence and significant computational burden. Another obstacle arises from the heterogeneous domain distributions (\eg, image styles, label definitions) across various scenarios. For example, as depicted in Fig.~\ref{Fig: teaser} (a), (b), and (c), the data can be collected in real scenes or synthesized virtually. Besides, comparing Fig.~\ref{Fig: teaser} (c) and (d), a class of objects may be annotated in a scene but not labeled in another scene, leading to optimization confusion during the network convergence process.

In this work, we address the challenges of unified 3D object detection from two perspectives, the algorithm perspective and data perspective. From the algorithm perspective, we develop a two-stage BEV detection architecture to tackle the different data geometry properties. In this architecture, the first stage produces initial target position estimation, and the second stage locates targets using this estimation as priori, which helps stabilize the convergence process. Moreover, we introduce an uneven BEV grid split strategy that expands the BEV space range while maintaining a manageable BEV grid size. A sparse BEV feature projection strategy is developed to reduce the projection computational cost by 82.6\%. In addition, to handle the heterogeneous domain distributions,  we propose a unified domain alignment technique consisting of two parts, the domain adaptive layer normalization (DALN) to align features, and the class alignment loss for alleviating label definition conflict. Combining all these innovative techniques, a well-behaved unified monocular 3D object detector is derived.

In terms of the data perspective, we observe that the unstable training of unified 3D object detectors is partly caused by the ill-posed nature of monocular depth estimation, and this nature can be effectively handled by incorporating depth sensors like lidars, which convert the monocular task into a multi-modal 3D object detection problem. Previous literature has advocated for camera-only detection due to the cost-effectiveness of cameras compared to depth sensors \cite{li2021monocular}.  Nevertheless, the prices of depth sensors are continuously decreasing and have now reached a level applicable for daily use. For instance, all Apple's iPads and iPhones have been equipped with lidars \cite{baruch2021arkitscenes}. Moreover, sparse point clouds are acceptable as deep learning models are capable of inferring depth based on nearby features \cite{wang2023lrru}. Therefore, inexpensive depth sensors present a viable option for multi-modal 3D perception, and we believe exploring unified multi-modal 3D object detection is valuable.

Nevertheless, despite the critical role of data in deep learning, there is no multi-modal 3D object detection benchmark that covers a diverse range of scenarios. To bridge this gap, we release the first multi-modal 3D object detection benchmark MM-Omni3D including multiple indoor and outdoor scenes. Several examples from MM-Omni3D are visualized in Fig.~\ref{Fig: teaser} (e)$\sim$(j), where significant variations in image styles and depth point densities can be observed. Each data pair in MM-Omni3D comprises an image and a point cloud, with a total of 159.5K pairs and 629.4K valid 3D box annotations. Afterwards, we extend the aforementioned monocular 3D object detector into its multi-modal version by adding a novel multi-modal feature fusion module named Mutual Information Collaboration (MIC), and it allows the image and point features to complement each other. This multi-modal detector serves as a simple baseline in MM-Omni3D.

In this work, we term the aforementioned monocular and multi-modal unified monocular object detectors as UniMODE and MM-UniMODE, respectively. Extensive experiments are conducted to validate the effectiveness of these two detectors and the developed composite benchmark MM-Omni3D. The results suggest that: (1) UniMODE achieves state-of-the-art (SOTA) performance on Omni3D, which is a large-scale popular monocular 3D object detection dataset. (2) The unstable training dynamics of unified detection can be resolved efficiently by incorporating depth information. (3) Co-training with data from more domains generally improves the detection accuracy and enhances the out-of-domain generalization capability. (4) In addition to serving as a multi-modal 3D object detection benchmark, MM-Omni3D also shows promise as a pre-training dataset.

This paper is an extended version of our conference paper \cite{li2024unimode}, where we make the following new contributions:

$\bullet$$\ $We build the first multi-modal 3D object detection benchmark that covers diverse scenarios, including both indoor and outdoor scenes. This benchmark enables us to study the unified 3D object detection problem not only in the monocular detection setting but also with the multi-modal detection setting. 

$\bullet$$\ $We design a novel multi-modal feature fusion module MIC and extend the monocular 3D object detector UniMODE to its multi-modal version, namely MM-UniMODE. This work serves as a pioneering effort to explore the unified multi-modal 3D object detection problem.

$\bullet$$\ $We draw several insightful findings that highlight the benefits of using multi-modal data through performing more extensive experiments. The experimental results indicate that employing multi-modal data for unified 3D object detection can efficiently address the unstable training problem. Besides, we reveal that MM-Omni3D can serve as an effective co-training or pre-training dataset.

\section{Related Work}
\label{Sec: Related Work}

\noindent \textbf{Monocular 3D object detection.} Due to the advantages of being economical and flexible, monocular 3D object detection grabs much research attention. Existing monocular 3D object detectors can be broadly categorized into two groups, camera-view detectors and BEV detectors. Among them, camera-view detectors first predict attributes that can be represented in the 2D image plane, such as projected 3D centers and depth. Then, the predicted attributes are converted into the 3D physical space based on camera intrinsic and extrinsic parameters \cite{li2022densely,peng2022did}, and the converted attributes can be used to generate 3D detection boxes. The network designs of camera-view detectors primarily follow the frameworks of popular 2D object detectors. For example, M3D-RPN \cite{brazil2019m3d} represents targets as 2D anchor boxes like Faster RCNN \cite{ren2016faster}. SMOKE \cite{liu2020smoke} utilizes the feature at the positions of the projected 3D centers to describe concerned objects, the idea of which is similar to CenterNet \cite{zhou2019objects}. A significant advantage of this camera-view detection paradigm is that the various attributes composing 3D detection boxes are disentangled, thereby easing the difficulty of analyzing results. However, the conversion from the 2D camera plane to 3D physical space can introduce additional errors \cite{wang2021fcos3d}, which negatively impact downstream planning tasks typically performed in 3D \cite{claussmann2019review}.

BEV detectors, on the other hand, transform image feature from the 2D camera plane to the 3D physical space before generating results in 3D \cite{li2023voxelformer}. This approach benefits downstream tasks, as planning is also performed in the 3D space \cite{reading2021categorical}. The 2D-to-3D feature transformation strategies adopted by existing BEV detectors can mainly be divided into three classes, \ie, projection based, sampling based, and implicit transformation based. In terms of the projection based strategies, they directly project the camera-view feature to the BEV grids pre-defined in the 3D space \cite{li2023bevdepth}. Differently, the sampling based strategies first project the 3D centers of BEV grids back to the camera view to sample feature, and then aggregate the sampled feature to derive BEV feature \cite{li2022bevformer}. For the implicit transformation based strategies, they fuse 2D image feature as 3D feature implicitly through global Transformer attention instead of explicit transformation \cite{liu2022petr}. Although the three strategies employ different implementations, their motivations are similar, \ie, directly generating detection boxes based on 3D feature. Nevertheless, these strategies also show limitations. All of them require accurate depth estimation (explicit or implicit), which can be challenging to achieve with only camera images \cite{park2021pseudo}. As a result, convergence could become unstable when dealing with diverse data scenarios \cite{brazil2023omni3d}.

\noindent \textbf{Multi-modal 3D object detection.} Although some previous detectors solely rely on cameras for cost-effectiveness, their performance is often compromised due to the ill-posed nature of monocular depth estimation \cite{liu2020smoke,wang2023exploring}. To handle this problem, there are works that exclusively use depth sensors, which yield precise 3D bounding boxes but may struggle with accurate category prediction due to the lack of semantic information in point clouds \cite{chen2023focalformer3d,zhan2023real}. As such, multi-modal 3D object detection, which combines cameras and depth sensors, presents a promising approach. The key research problem in multi-modal 3D object detection compared with monocular 3D object detection is how to fuse the multi-modal feature. To study this problem, early methods like EPNet \cite{huang2020epnet} decorate raw depth points with sampled image feature. However, when the input point cloud is sparse, a part of semantic information contained in the image feature could be discarded. There are also methods that utilize input depth points to assist image-based depth estimation and then produce more accurate 3D detection boxes \cite{qi2018frustum}. This paradigm of methods usually depends heavily on precise depth estimation, which is not always attainable because of the sparse input point cloud.  Recently, inspired by the BEV detection architectures, several methods first transform the image feature and depth feature into the same space and then fuse them \cite{liang2022bevfusion}. Compared with previous methods, this paradigm can fully exploit both image and depth information, and the dependence on accurate depth estimation is also eased. Notably, the space used for fusing feature could be implicit. For instance, CMT \cite{yan2023cross} encodes both image and point feature as tokens and fuses them through iterative Transformer attention based feature interaction. Notably, although there have been plentiful works about multi-modal 3D object detection, current detectors are limited to training and evaluation within a single data scenario, leading to these detectors specializing in a limited number of domains \cite{wang2021fcos3d}. While the modules adopted by these detectors improve the performance in some domains, they could negatively impact detection precision in others. To bridge this gap, this work expects to develop a multi-modal 3D object detector that is trained and evaluated across multiple data scenarios.

\noindent \textbf{Unified object detection.} In order to improve the generalization ability of detectors, some works have explored the integration of multiple data sources during model training \cite{wang2019towards,li2022efficient}. For example, in the field of 2D object detection, SMD \cite{zhou2022simple} observes that a significant challenge existing in unifying multiple domains of data to trained detectors is the label conflict between domains. To address the challenge, this work proposes to use the same detection network but different training protocols for each dataset. In this way, through scaling up the training data volume, a unified label space is learned and the detection accuracy is boosted. Similarly, in the 3D object detection domain, PPT~\cite{wu2023towards} investigates the utilization of extensive 3D point cloud data from diverse datasets for pre-training detectors. Unlike SMD, PPT aligns the knowledge from various domains through training a small set of domain-specific parameters. In addition, Uni3DETR~\cite{wang2023uni3detr} reveals how to devise a unified point-based 3D object detection architecture that can be applied to different datasets. However, an individual detector needs to be trained for each dataset, which means the network parameter is not shared. For the camera-based detection track, Cube RCNN~\cite{brazil2023omni3d} serves as the sole predecessor in the study of unified monocular 3D object detection. However, Cube RCNN is plagued by the unstable convergence issue, necessitating further in-depth analysis within this track.

\section{Unified Monocular Detection}
\label{Sec: Unified Monocular Detection}

In this section, we introduce our proposed unified monocular 3D object detector UniMODE. Specifically, in Section~\ref{SubSec: UniMODE Overall Framework}, we describe its overall framework. In Section~\ref{SubSec: Two-Stage Detection Architecture}, Section~\ref{SubSec: Uneven BEV Grid}, and Section~\ref{SubSec: Sparse BEV Feature Projection}, we elaborate on the developed two-stage detection architecture, uneven BEV grid, and sparse BEV feature projection strategy, respectively. These designs are to bridge the diverse geometry properties of various training domains. Then, in Section~\ref{SubSec: Unified Domain Alignment}, we explain the strategies utilized to address the heterogeneous domain distributions.

\begin{figure*}[tb]
    \centering
    \includegraphics[width=1.0\textwidth]{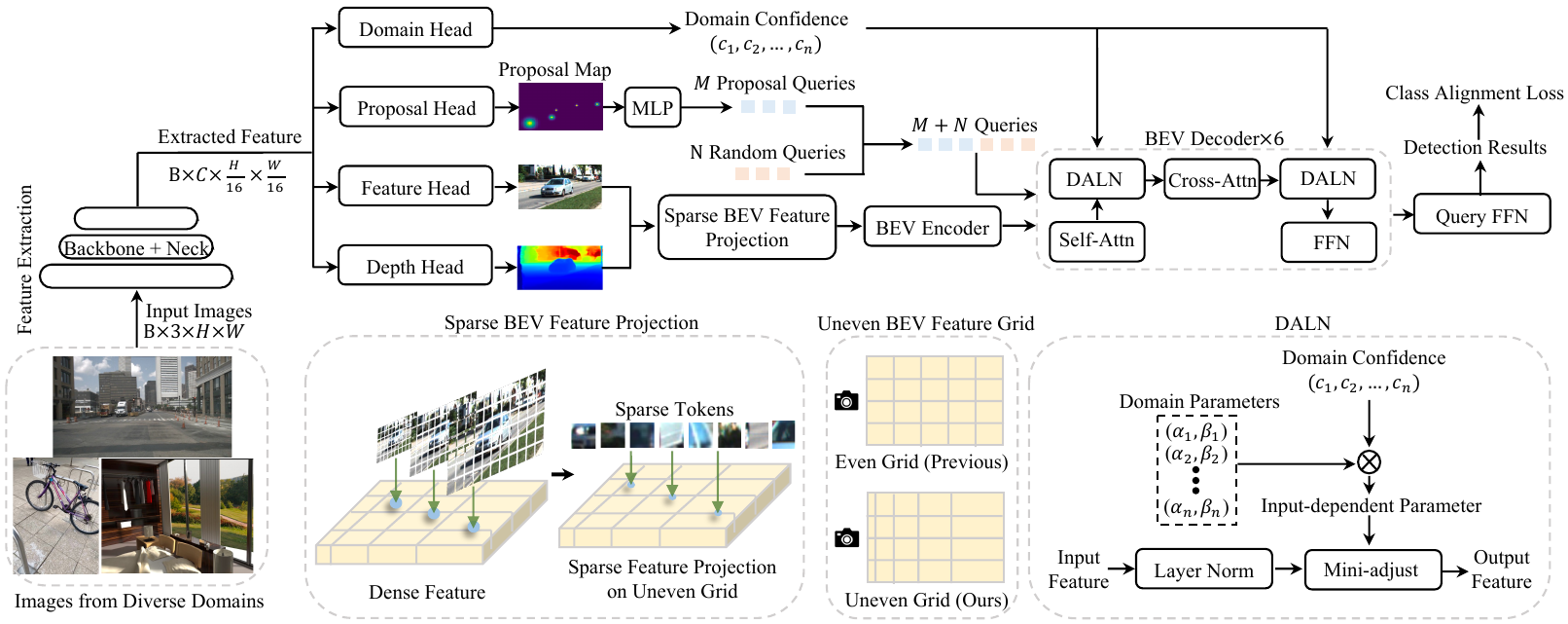}
    \caption{The overall detection framework of UniMODE. The illustrated modules proposed in this work include the proposal head, sparse BEV feature projection, uneven BEV feature grid, domain adaptive layer normalization, and class alignment loss.} \label{Fig: Pipeline}
\end{figure*}

\subsection{UniMODE Overall Framework}
\label{SubSec: UniMODE Overall Framework}

The overall framework of UniMODE is illustrated in Fig.~\ref{Fig: Pipeline}. As shown, a monocular image $I \in \mathbb{R}^{3 \times H \times W}$ sampled from multiple scenarios (\eg, indoor and outdoor, real and synthetic, daytime and nighttime) are input to the feature extraction module (including a backbone and a neck) to produce representative feature $F \in \mathbb{R}^{C \times \frac{H}{16} \times \frac{W}{16}}$. Then, $F$ is processed by 4 fully convolutional heads, namely ``domain head", ``proposal head", ``feature head", and ``depth head", respectively. Among them, the role of the domain head is to predict which pre-defined data domain an input image is most relevant to, and the classification confidence produced by the domain head is subsequently utilized in domain alignment. The proposal head aims to estimate the rough target distribution before the 6 Transformer decoders, and the estimated distribution serves as prior information for the second-stage detection. This design alleviates the distribution mismatch between diverse training domains (refer to Section~\ref{SubSec: Two-Stage Detection Architecture}). The proposal head output is encoded as $M$ proposal queries. In addition, $N$ queries are randomly initialized and concatenated with the proposal queries for the second-stage detection, leading to $M+N$ queries in the second stage. 

The feature head and depth head are responsible for projecting the image feature into the BEV plane and obtaining the BEV feature. During this projection, we develop a technique to remove unnecessary projection points, which reduces the computing burden by about 82.6\% (refer to Section~\ref{SubSec: Sparse BEV Feature Projection}). Besides, we propose the uneven BEV feature (refer to Section~\ref{SubSec: Uneven BEV Grid}), which means the BEV grids closer to the camera enjoy more precise resolution, and the grids farther to the camera cover broader perception areas. This design well balances the grid size contradiction between indoor detection and outdoor detection without extra memory burden.

Obtaining the projected BEV feature, a BEV encoder is employed to further refine the feature, and 6 decoders are adopted to generate the second-stage detection results. As mentioned before, $M+N$ queries are used during this process. After the 6 decoders, the queries are decoded as detection results by querying the FFN. In the decoder part, the unified domain alignment strategy is devised to align the data of various scenarios via both the feature and loss perspectives. Refer to Section~\ref{SubSec: Unified Domain Alignment} for more details. 

\subsection{Two-Stage Detection Architecture}
\label{SubSec: Two-Stage Detection Architecture}
\begin{figure}[tbp]
    \centering
    \includegraphics[width=1.0\linewidth]{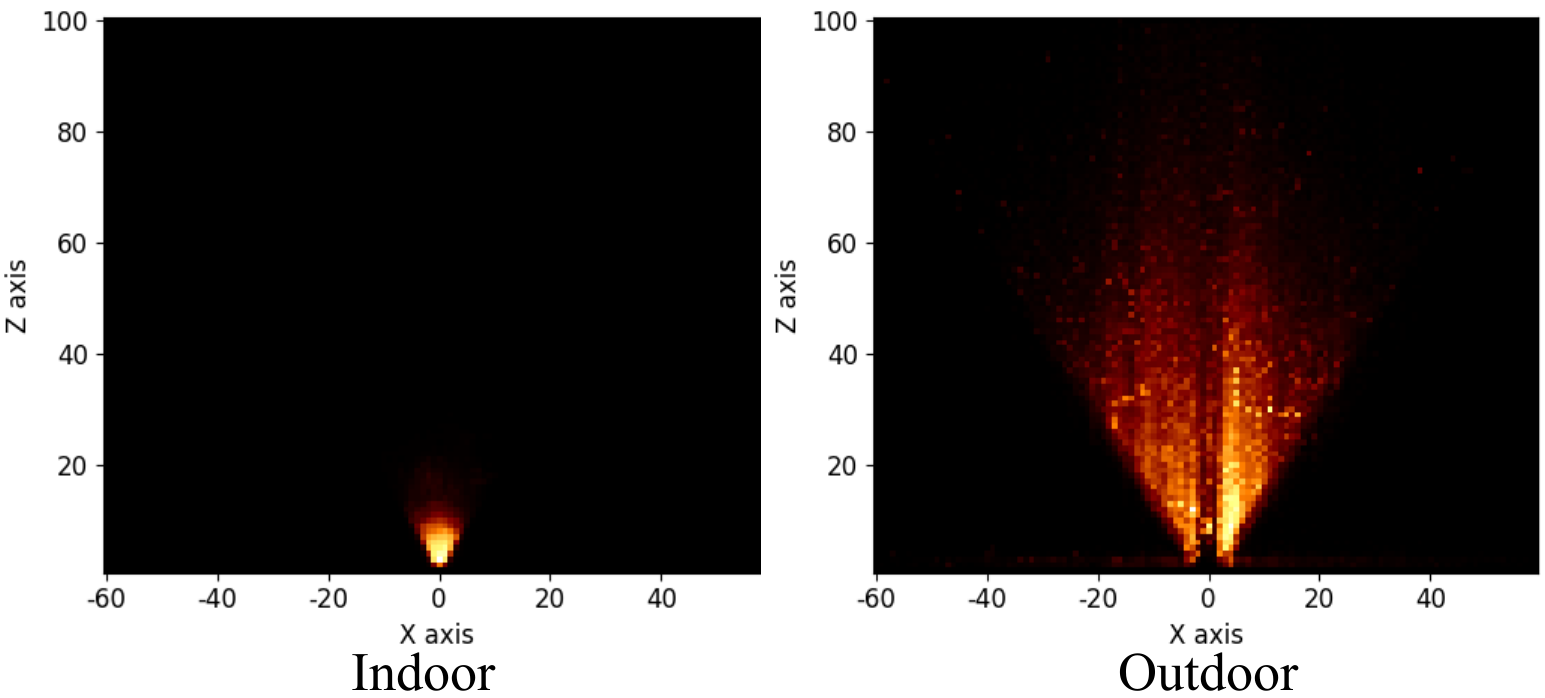}
    \caption{Indoor and outdoor target position distributions in the BEV space. The brighter a point shows, the more targets the corresponding BEV grid contains. The perception camera is located at the point with the coordinate $(0, 0)$.} \label{Fig: data_statistics}
\end{figure}

The integration of indoor and outdoor 3D object detection is challenging due to diverse geometry properties (\eg, perception ranges, target positions). Indoor detection typically involves close-range targets, while outdoor detection concerns targets scattered over a broader 3D space. As depicted in Fig.~\ref{Fig: data_statistics}, the perception ranges and target positions in indoor and outdoor detection scenes vary significantly, which are challenging for traditional BEV 3D object detectors because of their fixed BEV feature resolutions.

The geometry property difference is identified as an essential reason causing the unstable convergence of BEV detectors \cite{li2022bevformer}. For example, the difference in target position distribution makes it challenging for Transformer-based detectors to learn how to update the query reference points gradually towards the concerned objects. In fact, through visualization, we find the reference point updating in the 6 Transformer decoders is disordered. As a result, if we adopt the classical deformable DETR architecture \cite{zhu2020deformable} to build a 3D object detector, the training is easy to collapse due to the inaccurate positions of learned reference points, resulting in sudden gradient vanishing or exploding.

To overcome this challenge, we construct UniMODE in a two-stage detection fashion. In the first stage, we design a CenterNet~\cite{zhou2019objects} style head (the proposal head in Fig.~\ref{Fig: Pipeline}) to produce detection proposals. Specifically, its predicted attributes include the 2D center Gaussian heatmap, offset from 2D centers to 3D centers, and 3D center depths of targets. The 3D center coordinates of proposals can be derived from these predicted attributes. Then, the proposals with top $M$ confidences are selected and encoded as $M$ proposal queries by an MLP layer. To account for any potential missed targets, another $N$ randomly initialized queries are concatenated with these proposal queries to perform information interaction in the 6 decoders of the second stage (the Transformer stage). In this way, the initial query reference points of the second detection stage are adjusted adaptively. Our experiments reveal that this two-stage architecture is essential for stable convergence.

Besides, since the positions of query reference points are not randomly initialized, the iterative bounding box refinement strategy proposed in deformable DETR \cite{zhu2020deformable} is abandoned as it may lead to a deterioration of reference point quality. In fact, we observe that this iterative bounding box refinement strategy could result in convergence collapse.

\subsection{Uneven BEV Grid}
\label{SubSec: Uneven BEV Grid}

A notable difference between indoor and outdoor 3D object detection lies in the geometry information (\eg, scale, proximity) of objects to the camera during data collection. Indoor environments typically feature smaller objects located closer to the camera, whereas outdoor environments involve larger objects positioned at greater distances. Furthermore, outdoor 3D object detectors must account for a wider perceptual range of the environment. Consequently, existing indoor 3D object detectors typically use smaller voxel or pillar sizes. For instance, the voxel size of CAGroup3D \cite{wang2022cagroup3d}, a SOTA indoor 3D object detector, is 0.04 meters, and the maximum target depth in the SUN-RGBD dataset \cite{song2015sun}, a classic indoor dataset, is approximately 8 meters. In contrast, outdoor datasets exhibit much larger perception ranges. For example, the commonly used outdoor detection dataset KITTI \cite{geiger2012we} has a maximum depth range of 100 meters. Due to this vast perception range and limited computing resources, outdoor detectors employ larger BEV grid sizes, \eg, the BEV grid size in BEVDepth \cite{li2023bevdepth}, a state-of-the-art outdoor 3D object detector, is 0.8 meters.

Therefore, the BEV grid sizes of current outdoor detectors are typically large to accommodate the vast perception range, while those of indoor detectors are small because of the intricate indoor scenes. However, since UniMODE aims to address both indoor and outdoor 3D object detection using a unified model structure and network weight, its BEV feature must cover a large perception area while still utilizing small BEV grids, which poses a massive challenge due to the limited GPU memory. To overcome this challenge, we propose a solution that involves partitioning the BEV space into uneven grids, in contrast to the even grids utilized by existing detectors. As depicted in the bottom part of Fig.~\ref{Fig: Pipeline}, we achieve this by employing a smaller size of grids closer to the camera and larger grids for those farther away. This approach enables UniMODE to effectively perceive a wide range of objects while maintaining small grid sizes for objects in close proximity. Importantly, this does not increase the total number of grids, thereby avoiding any additional computational burden. Specifically, assuming there are $N_z$ grids in the depth axis and the depth range is $(z_{min}, z_{max})$, the grid size of the $i_{\rm th}$ grid $z_{i}$ is set to:
\begin{align}
z_{i} = z_{min} + \frac{z_{max} - z_{min}}{N_{z}(N_{z} + 1)} \cdot i(i + 1). \label{Eq1}
\end{align}

Notably, the mathematical form in Eq.~\ref{Eq1} is similar to the linear-increasing discretization of depth bin in CaDDN \cite{reading2021categorical}, while the essence is fundamentally different. In CaDDN, the feature projection distribution is adjusted to allocate more features to the grids that are closer to the camera. In experiments, we observe that this adjustment results in a more imbalanced BEV feature, \ie, denser features in closer grids and more empty grids in farther grids. Since features in all grids are extracted by the same network, this imbalance degrades the 3D object detection performance. By contrast, our uneven BEV grid approach enhances detection precision by making the feature density more balanced.

\subsection{Sparse BEV Feature Projection}
\label{SubSec: Sparse BEV Feature Projection}

The step of transforming the camera view feature into the BEV space is quite computationally expensive due to its numerous projection points. Specifically, considering the image feature $F_{i} \in \mathbb{R}^{C_i \times 1 \times H_f \times W_f}$ and depth feature $F_{d} \in \mathbb{R}^{1 \times C_d \times H_f \times W_f}$, the projection feature $F_{p} \in \mathbb{R}^{C_i \times C_d \times H_f \times W_f}$ is obtained by multiplying $F_{i}$ and $F_{d}$. Therefore, the projection point number $C_i \times C_d \times H_f \times W_f$ increases dramatically as the growth of $C_{d}$. The heavy computational burden of this feature projection step restricts the BEV feature resolution, and thus hinders unifying indoor and outdoor 3D object detection. In this work, we observe that most projection points in $F_p$ are unnecessary because their values are quite tiny. This is essentially because of the small corresponding values in $F_{d}$, which imply that the model predicts there is no target in these specific BEV grids. Hence, the time spent on projecting features to these unconcerned grids can be saved.

Based on the above insights, we propose to remove the unnecessary projection points based on a pre-defined threshold $\tau$. Specifically, we eliminate the projection points in $F_p$ whose corresponding depth confidence of $F_d$ is smaller than the threshold $\tau$. In this way, most projection points are eliminated. For instance, when setting $\tau$ to 0.001, about 82.6\% of projection points can be excluded.

\subsection{Unified Domain Alignment}
\label{SubSec: Unified Domain Alignment}
%In this work, we address the difference between data scenarios from two views, the feature view and loss view.
Heterogeneous domain distributions exist in diverse scenarios and we address this challenge from the feature view and loss view separately.

\noindent \textbf{Domain adaptive layer normalization.} For the feature view, we initialize domain-specific learnable parameters to address the variations observed in diverse training data domains. However, this strategy must adhere to two crucial requirements. Firstly, the detector should exhibit robust performance during inference, even when confronted with images from domains that are not encountered during training. Secondly, the introduction of these domain-specific parameters should incur minimal computational overhead.

Considering these two requirements, we propose the domain adaptive layer normalization (DALN) strategy. In this strategy, we first split the training data into $D$ domains. For the classic implementation of layer normalization (LN) \cite{ba2016layer}, denoting the input sequence as $X_l \in \mathbb{R}^{B \times L \times C}$ and its element with the index $(b, l, c)$ as $x_{l}^{(b, l, c)}$, the corresponding output $\hat{x}_{l}^{(b, l, c)}$ of processing $x_{l}^{(b, l, c)}$ by LN is obtained as:
\begin{align}
\hat{x}_{l}^{(b, l, c)} = \frac{x_{l}^{(b, l, c)} - \mu^{(b, l)}}{\sigma^{(b, l)}}, \label{Eq2}
\end{align}
where
\begin{align}
\mu^{(b, l)} = \frac{1}{C} \sum\limits_{i=1}^{C} x_{l}^{(b, l, i)}, 
\sigma^{(b, l)} = \sqrt{\frac{1}{C} \sum\limits_{i=1}^{C}( x_{l}^{(b, l, c)} - \mu^{(b, l)} )^{2}}. \label{Eq3}
\end{align}

In DALN, we build a set of learnable domain-specific parameters, \ie, $\{(\alpha_i, \beta_i)\}_{i=1}^{D}$, where $(\alpha_i, \beta_i)$ are the parameters corresponding to the $i_{\rm th}$ domain. $\{\alpha_i\}_{i=1}^{D}$ are initialized as 1 and $\{\beta_i\}_{i=1}^{D}$ are set to 0. Then, we establish a domain head consisting of several convolutional layers. As shown in Fig.~\ref{Fig: Pipeline}, the domain head takes the feature $F$ as input and predicts the confidence scores that the input images $I$ belong to these $D$ domains. Denoting the confidence of the $b_{\rm th}$ image as $\{c_{i}\}_{i=1}^{D}$, the input-dependent parameters $(\alpha, \beta)$ are computed following:
\begin{align}
\alpha = \sum\limits_{i=1}^{D} c_{i} \cdot \alpha_{i}, \; \beta = \sum\limits_{i=1}^{D} c_{i} \cdot \beta_{i}. \label{Eq4}
\end{align}
Obtaining $(\alpha, \beta)$, we employ them to adjust the distribution of $\hat{x}_{l}^{(b, l, c)}$ with respect to $\bar{x}_{l}^{(b, l, c)} = \alpha \cdot \hat{x}_{l}^{(b, l, c)} + \beta$, where $\bar{x}_{l}^{(b, l, c)}$ denotes the updated distribution. In this way, the feature distribution in UniMODE can be adjusted according to the input images self-adaptively, and the increased learnable parameters are negligible. Additionally, when an image unseen in the training set is input to the detector, DALN still works well, because the unseen image can still be classified as a weighted combination of these $D$ domains.

Although there exist a few previous techniques related to adaptive normalization, almost all of them are based on regressing input-dependent parameters directly \cite{wu2023towards}. So, they need to build a special regression head for every normalization layer. By contrast, DALN enables all layers to share the same domain head, so the computing burden is much smaller. Besides, DALN introduces domain-specific parameters, which are more stable to train.

\begin{figure}[tbp]
    \centering
    \includegraphics[width=1.0\linewidth]{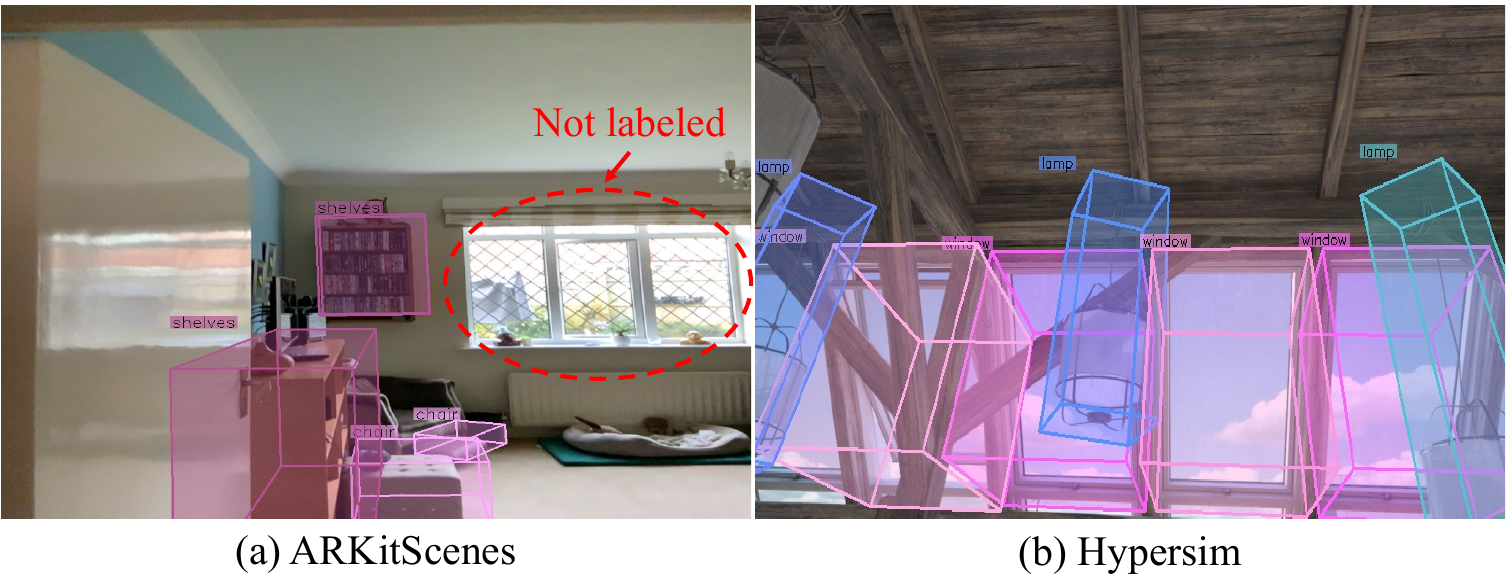}
    \caption{An example of heterogeneous label conflict among sub-datasets in Omni3D. As shown, ``Window'' is not labeled in ARKitScenes while labeled in Hypersim, so the unlabeled window in (a) could harm the convergence stability of detectors.} \label{Fig: class_conflict}
\end{figure}

\noindent \textbf{Class alignment loss.} In the loss view, we aim to address the heterogeneous label conflict when combining multiple data sources. Specifically, there are 6 independently labeled sub-datasets in Omni3D, and their label spaces are different. For example, as presented in Fig.~\ref{Fig: class_conflict}, although the \textit{Window} class is annotated in ARKitScenes, it is not labeled in Hypersim. As the label space of Omni3D is the union of all classes in all subsets, the unlabeled window in Fig.~\ref{Fig: class_conflict} (a) becomes a missing target that harms convergence stability.

The two-stage detection architecture described in Section~\ref{SubSec: Two-Stage Detection Architecture} can alleviate the aforementioned problem to some extent, because it helps the detector concentrate on foreground objects, and the unlabeled objects are overlooked to compute loss. To address this problem further, we devise a simple strategy, i.e., the class alignment loss. Specifically, denoting the label space of the $i_{\rm th}$ dataset as $\Omega_{i}$, we compute loss on the $i_{\rm th}$ dataset as:
\begin{align}
L_{i} = \{
\begin{array}{lcl}
\gamma \cdot l(y, \bar{y}), (\bar{y} \notin \Omega_{i}) \wedge (\bar{y} = \mathcal{B}) \\
l(y, \bar{y}), others \\
\end{array}
, \label{Eq5}
\end{align}
where $l(\cdot)$, $y$, $\bar{y}$, $\mathcal{B}$ are the loss function, class prediction, class label, and background class, respectively. $\gamma$ is a factor for reducing the punishment to the classes that are not included in the label space of this sample.

\section{Unified Multi-modal Detection}
\label{SubSec: Unified Multi-modal Detection}

In this section, we describe how we achieve unified multi-modal 3D object detection. Specifically, due to the lack of a proper benchmark, we first explain how we construct the unified multi-modal 3D object detection benchmark MM-Omni3D in Section~\ref{SubSec: MM-Omni3D Benchmark}. Then, we present the data statistics of the data in this benchmark in Section~\ref{SubSec: MM-Omni3D Data Statistics}. In Section~\ref{SubSec: MM-UniMODE Overall Framework}, we elaborate on the overall framework of the developed multi-modal detector MM-UniMODE, which is extended from the aforementioned UniMODE. Finally, the designed multi-modal feature fusion strategy MIC is introduced in Section~\ref{SubSec: Mutual Information Collaboration}.

\subsection{MM-Omni3D Benchmark}
\label{SubSec: MM-Omni3D Benchmark}

MM-Omni3D consists of six datasets, \ie, SUN-RGBD \cite{song2015sun}, ARKitScenes \cite{baruch2021arkitscenes}, Objectron \cite{ahmadyan2021objectron}, Hypersim \cite{roberts2021hypersim}, KITTI \cite{geiger2012we}, and nuScenes \cite{caesar2020nuscenes}. Among them, SUN-RGBD, ARKitScenes, Objectron, and Hypersim are indoor datasets, while KITTI and nuScenes are outdoor datasets. Besides, Hypersim is a synthesized dataset and the other five datasets are collected from real scenes by different sensors.

A multi-modal dataset needs to incorporate at least one depth sensor in addition to a camera. This depth sensor can be of various types, including lidar, radar, RGBD camera, time-of-flight camera, structure from motion, etc. Previous datasets usually use the same depth sensor across the whole dataset \cite{cippitelli2017radar}. Therefore, the depth point distributions in these datasets are highly similar, and the trained detectors often only perform well when they are tested in scenes that employ the same type of depth sensor. To satisfy the diverse requirements of practical applications, we argue that a multi-modal 3D object detector should utilize camera images and point clouds from various kinds of depth sensors. This is because training a high-performance detector consumes a lot of data. However, collecting multi-modal data is expensive. It is often impractical to prepare abundant multi-modal training data in every specific scene. To handle this challenge, training a universal detector by combining data from various available scenes is feasible.

\begin{figure}[tbp]
    \centering
    \includegraphics[width=1.0\linewidth]{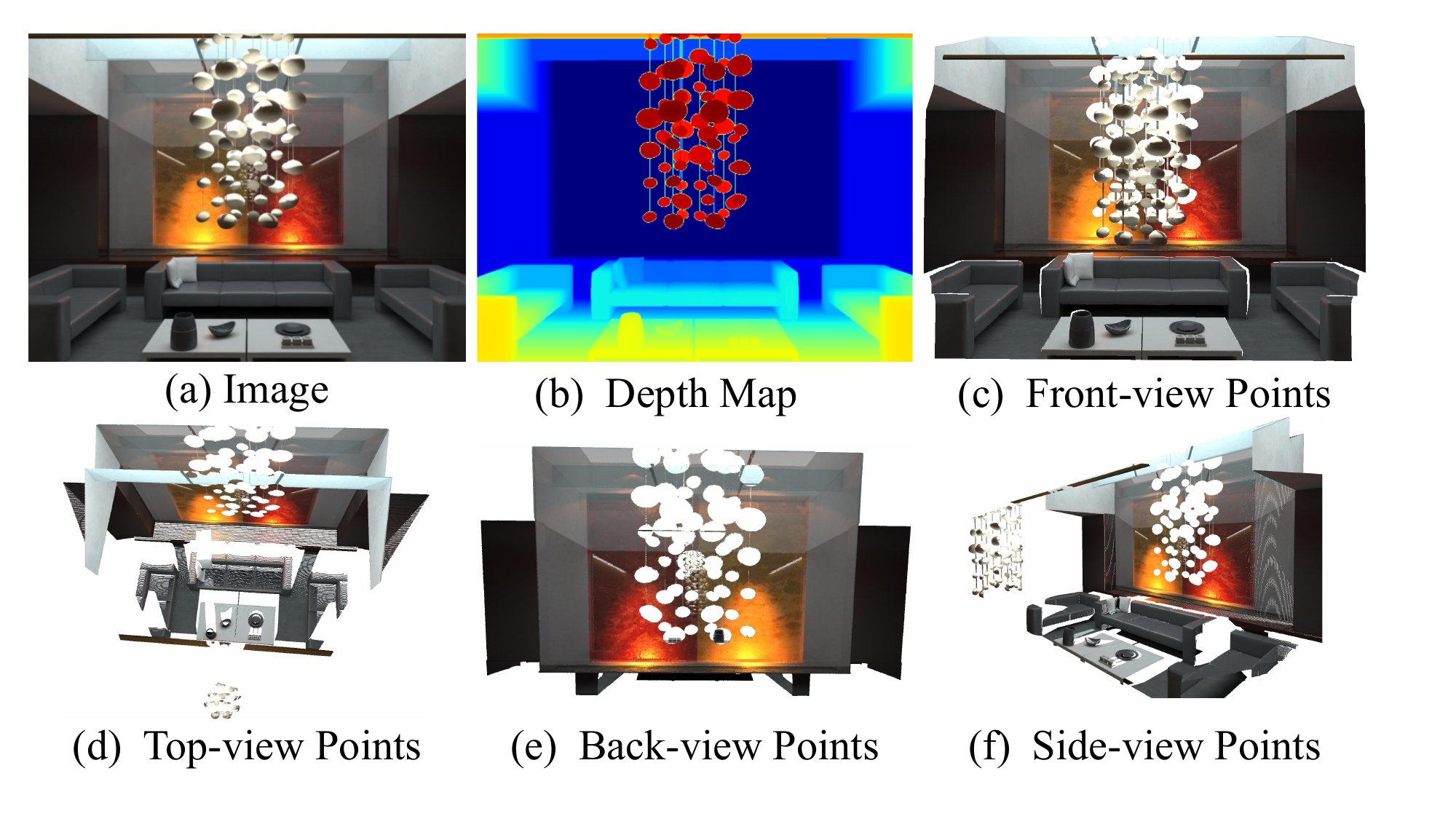}
    \caption{(a) Camera-view image. (b) Camera-view depth map. (c)$\sim$(f) The processed point clouds observed from the front view, top view, back view, and side view, respectively. As shown, unlike previous 3D object detection datasets \cite{song2015sun} that store the point cloud of the whole scene, MM-Omni3D only provides the points visible from the camera view, which better meets the practical situation of online multi-modal 3D object detection.} \label{Fig: camera_view_depth}
\end{figure}

Nevertheless, the depth from various sensors differs not only in data formats but also in the employed 3D coordinate systems. For example, the depth could be a depth map \cite{geiger2012we} as shown in Fig.~\ref{Fig: camera_view_depth} (b) or point cloud \cite{caesar2020nuscenes} in Fig.~\ref{Fig: camera_view_depth} (c). To address this problem, we transform all depth information into a unified point cloud format and the same camera-view coordinate system. Given that 3D object detection is an online perception task, the multi-modal detector should only have access to depth points that are directly observable by the perception sensors. However, in some data samples, the point cloud is collected in an offline manner, meaning the point cloud for the entire scene is gathered \cite{song2015sun}. To solve this problem, as illustrated in Fig.~\ref{Fig: camera_view_depth} (c)$\sim$(f), we remove all depth points invisible to the camera view at the current timestamp and only retain the visible points. These visible point clouds are associated with image pixels based on camera parameters in a unified format. After investing significant effort in unifying the data formats of various datasets and evaluation procedures, we obtain the MM-Omni3D benchmark. The evaluation metrics used in this benchmark are the same as Omni3D \cite{brazil2023omni3d}.

\begin{figure}[tbp]
    \centering
    \includegraphics[width=1.0\linewidth]{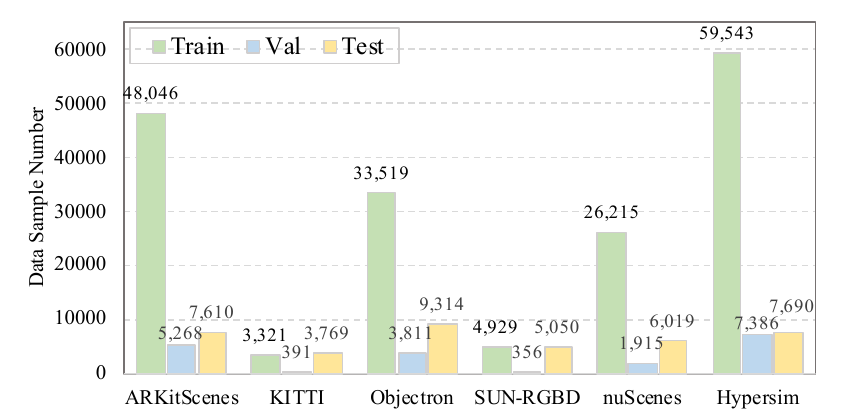}
    \caption{The data sample numbers of the train, validation, and test sets in the six datasets incorporated in MM-Omni3D.} \label{Fig: data_sample_number}
\end{figure}

\subsection{MM-Omni3D Data Statistics}
\label{SubSec: MM-Omni3D Data Statistics}

The MM-Omni3D dataset consists of a total of 234,152 data samples and is divided into 6 subsets, each corresponding to a different data scenario as shown in Fig.~\ref{Fig: teaser}~(e)$\sim$(j). Among these subsets, four are dedicated to indoor object detection, while the remaining two pertain to outdoor scenes. Each subset is further divided into training, validation, and testing sets. The number of samples in these subsets is illustrated as Fig.~\ref{Fig: data_sample_number}, Specifically, for the entire MM-Omni3D dataset, the data sample numbers of the training, validation, and testing sets are about 175.6K, 19.1K, and 39.5K, respectively. In MM-Omni3D, a total of 50 object classes are annotated. It is important to note that only valid 3D box annotations are considered, where the projected 3D centers are visible in the camera images. In addition, the distribution of instances across classes presents a long tail, indicating a significant imbalance in the instance numbers of different classes. For example, there are 152K \textit{Car} instances in MM-Omni3D while only 21 \textit{Floor mat} instances, which suggests that the head class with the most instances is about 7.2K times more than the tail class.

The adopted depth sensors are also diverse in MM-Omni3D. As described in Fig.~\ref{Fig: teaser}~(f)$\sim$(g), although both KITTI and nuScenes employ lidars, their ray numbers are different, resulting in significant variations in point densities. Similarly, the depth points in Objectron and ARKitScenes are obtained through AR sessions, but their densities are quite different due to the discrepancies in collecting data procedures and post-processing operations. In contrast to the aforementioned datasets, the depth points in SUN-RGBD are collected by multiple types of RGB-D cameras. To the best of our knowledge, MM-Omni3D is the first dataset to include such a wide range of depth sensors. The position distributions of targets in MM-Omni3D are also diverse. It can be observed from Fig.~\ref{Fig: teaser}~(e)$\sim$(j) that indoor targets are mostly in close proximity to the camera, while outdoor objects are far away. This distribution difference could lead to optimization conflict and training instability.

\begin{figure*}[tb]
    \centering
    \includegraphics[width=1.0\textwidth]{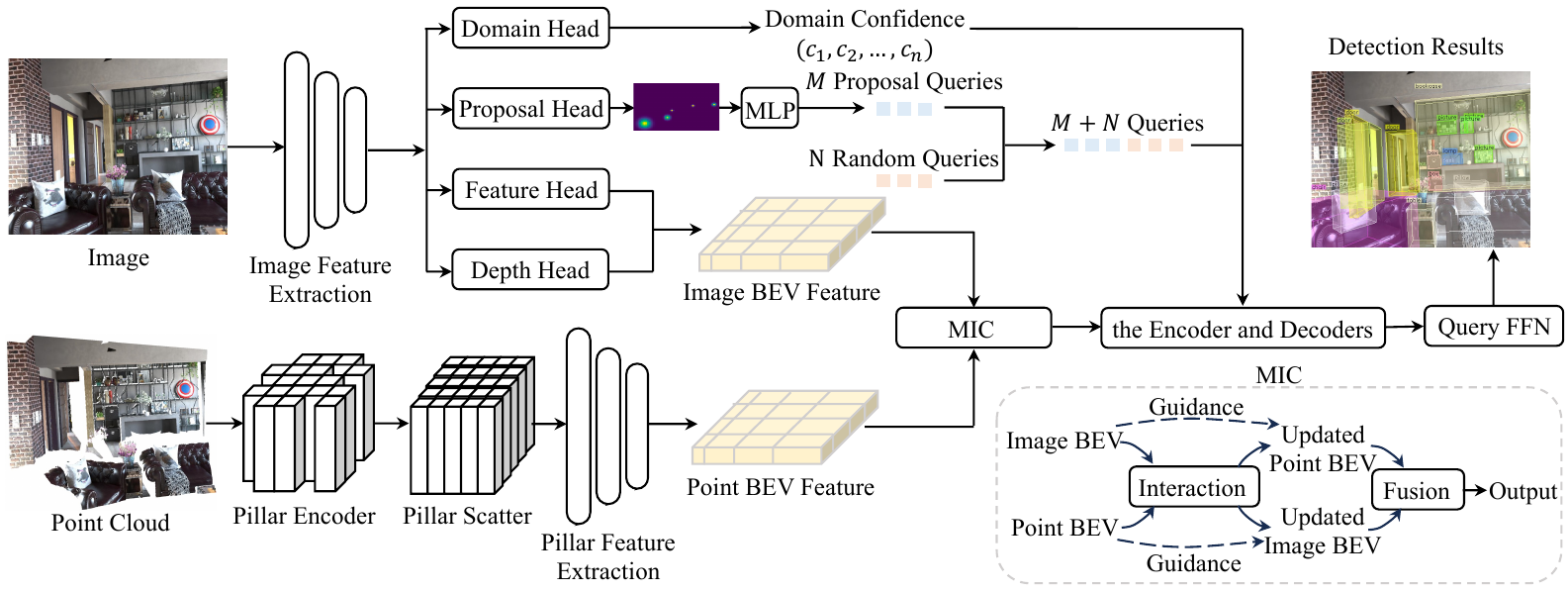}
    \caption{The overall detection framework of MM-UniMODE. For the clarity of this figure, we omit the details of modules that have been illustrated in Fig.~\ref{Fig: Pipeline}. The newly added modules include the point cloud feature extraction branch and the MIC for image-point feature fusion.} \label{Fig: MM_Pipeline}
\end{figure*}

\subsection{MM-UniMODE Overall Framework}
\label{SubSec: MM-UniMODE Overall Framework}

In this part, we elaborate on the overall framework of the unified multi-modal 3D object detector MM-UniMODE, which is extended from the aforementioned detector UniMODE. The pipeline of MM-UniMODE is illustrated as Fig.~\ref{Fig: MM_Pipeline}. As shown, the detector takes an image and a point cloud as input. The image is captured by a camera, and the point cloud can be collected by any kind of depth sensor. An image branch and a point branch are constructed to extract features from the image and point cloud separately.

The image branch of MM-UniMODE is the same as UniMODE, and its output includes the image BEV feature, domain confidence, and $M+N$ queries.  For the point branch, the input point cloud is first voxelized into 3D voxels and then processed by a pillar encoder employed in SECOND \cite{yan2018second}. Next, after the pillar scatter operation that transforms the obtained feature as a BEV feature, a point network (including a point backbone and a point neck) is employed to generate the point BEV feature. Obtaining the image and point BEV feature, we fuse them with the proposed MIC module, which enables the image BEV feature and point BEV feature to complement the information of each other, and then the multi-modal BEV feature is derived. Subsequently, 1 encoder and 6 decoders with the same structure implementation as UniMODE are employed to generate detection results based on the multi-modal BEV feature.

\subsection{Mutual Information Collaboration}
\label{SubSec: Mutual Information Collaboration}

For every training sample, MM-Omni3D takes an image and a point cloud as input and produces the image BEV feature $B_{I}$ and point BEV feature $B_{P}$. Due to the data modality difference, $B_{I}$ and $B_{P}$ present significantly different characteristics. For example, as shown in Fig.~\ref{Fig: data_statistics}, the indoor points are denser and closer to the ego sensors, while the outdoor points are sparser and farther. Since the detector needs to detect both indoor and outdoor targets, the size of the BEV feature has to cover the biggest required perception range in MM-Omni3D. Consequently, in outdoor scenes, a large part of $B_{P}$ could do not contain any depth point, and the learned feature in these positions is invalid. Meanwhile, for $B_{I}$, since monocular depth estimation is an ill-posed problem, the extracted feature may be projected to a false grid, which results in the ambiguity in $B_{I}$. Therefore, both the image and point branches have their limitations. Thus we propose the MIC module to fuse image-point BEV features and address the limitations of two branches by enabling these two branches to learn from each other collaboratively, the basic concept of which is illustrated Fig.~\ref{Fig: MM_Pipeline}.

Specifically, given $B_{I}$ and $B_{P}$, we first conduct information interaction between them to obtain the updated BEV feature $\hat{B}_{I}$ and $\hat{B}_{P}$. The interaction is implemented using two convolutional layers. For the image feature, the main problem is the quality of $\hat{B}_{I}$ is unsatisfactory due to imprecise depth estimation. By contrast, the feature in grids of $B_{P}$ that contain points is informative. Therefore, we first identify which grids in $B_{P}$ have depth points and denote the boolean mask that suggests the positions of these grids as $M_{P}$. Then, the feature of these grids in $B_{P}$ and $\hat{B}_{I}$ can be represented as $B_{P} M_{P}$ and $\hat{B}_{I} M_{P}$, respectively. Then, we use $B_{P} M_{P}$ to provide guidance to $\hat{B}_{I} M_{P}$ by computing a guidance loss as:
\begin{align}
L_{P \rightarrow I} = l_{1} ( \mathcal{S}( B_{P} M_{P} ), \hat{B}_{I} M_{P} ), \label{Eq6}
\end{align}
where $l_{1}$ is the mean absolute error loss and $\mathcal{S}(\cdot)$ denotes stopping gradient.

In the point branch, the limitation is that numerous BEV grids exactly contain no depth point, so the learned feature in these grids is unreliable. By contrast, most grids in $B_{I}$ contain meaningful features because they are generated based on an estimated depth distribution \cite{reading2021categorical}. Therefore, we propose to employ $B_{I}$ to guide $\hat{B}_{P}$. Denoting the boolean mask that indicates the positions of grids with depth probabilities bigger than a hyper-parameter $\epsilon$ as $M_{I}$, the guidance loss is formulated as:
\begin{align}
L_{I \rightarrow P} = l_{1} (\hat{B}_{P} M_{I} (1-M_{P}), \mathcal{S}(B_{I} M_{I} (1-M_{P})), \label{Eq7}
\end{align}
where $M_{I} (1-M_{P})$ exactly suggests the positions of grids where the image feature is informative but the point feature is unreliable. Obtaining $\hat{B}_{I}$ and $\hat{B}_{P}$, we fuse them as the multi-modal BEV feature using a convolutional block. The losses $L_{P \rightarrow I}$ and $L_{I \rightarrow P}$ described in Eq.~\ref{Eq6} and Eq.~\ref{Eq7} are only utilized during training and not applied to inference.

\section{Experiment}
\label{Sec: Experiment}

In this section, we conduct extensive experiments to validate the effectiveness of the proposed modules used in UniMODE and MM-UniMODE. Additionally, we also study their out-of-domain generalization characteristics and the benefits of the developed multi-modal 3D object detection benchmark MM-Omni3D on serving as a pre-training dataset. To explain the experimental details clearly, we first present some basic information about these experiments as follows:

\noindent \textbf{Implementation details.} The perception ranges in the X-axis, Y-axis, and Z-axis of the camera coordinate system are $(-30, 30)$, $(-40, 40)$, $(0, 80)$ meters, respectively. If without a special statement, the BEV grid resolution is $(60, 80)$. The factor $\gamma$ defined in the class alignment loss is set to 0.2. The parameter $\epsilon$ for the MIC module is $5e^{-4}$. $M$ and $N$ are set to 100. The adopted optimizer is AdamW, and the learning rate is set to $12e^{-4}$ for a batch size of 192. Both UniMODE and MM-UniMODE are optimized for a total of 116k training iterations. The experiments in Table~\ref{Table: Monocular Performance Comparison} and Table~\ref{Table: Detailed Performance} are performed using 16 A100 GPUs and the remaining experiments are primarily conducted with 4 A100 GPUs.

\begin{table*}[tbp] 
    \centering
    \caption{Performance comparison between UniMODE and other monocular 3D object detectors.} \label{Table: Monocular Performance Comparison}
    \vspace{-0.2cm}
    \resizebox{1.0\linewidth}{!}{
    \begin{tabular}{c|ccc|cc|cccccc}
    \toprule
    \multirow{2}{*}{Method} & \multicolumn{3}{c|}{\rm OMNI3D\_{OUT}} & \multicolumn{2}{c|}{\rm OMNI3D\_{IN}} & \multicolumn{6}{c}{\rm OMNI3D} \\
    & ${\rm AP^{kit}_{3D}} \uparrow$ 
    & ${\rm AP^{nus}_{3D}} \uparrow$ 
    & ${\rm AP^{out}_{3D}} \uparrow$ 
    & ${\rm AP^{sun}_{3D}} \uparrow$ 
    & ${\rm AP^{in}_{3D}} \uparrow$ 
    & ${\rm AP_{3D}^{25}} \uparrow$ 
    & ${\rm AP_{3D}^{50}} \uparrow$
    & ${\rm AP_{3D}^{near}} \uparrow$ 
    & ${\rm AP_{3D}^{med}} \uparrow$ 
    & ${\rm AP_{3D}^{far}} \uparrow$ 
    & ${\rm AP_{3D}} \uparrow$ \\
    \midrule
    M3D-RPN \cite{brazil2019m3d} & 10.4\% & 17.9\% & 13.7\% & - & - & - & - & - & - & - & - \\
    SMOKE \cite{liu2020smoke} & 25.4\% & 20.4\% & 19.5\% & - & - & - & - & - & - & - & 9.6\% \\
    FCOS3D \cite{wang2021fcos3d} & 14.6\% & 20.9\% & 17.6\% & - & - & - & - & - & - & - & 9.8\% \\
    PGD \cite{wang2022probabilistic} & 21.4\% & 26.3\% & 22.9\% & - & - & - & - & - & - & - & 11.2\% \\
    GUPNet \cite{lu2021geometry} & 24.5\% & 20.5\% & 19.9\% & - & - & - & - & - & - & - & - \\
    ImVoxelNet \cite{rukhovich2022imvoxelnet} & 23.5\% & 23.4\% & 21.5\% & 30.6\% & - & - & - & - & - & - & 9.4\% \\
    BEVFormer \cite{li2022bevformer} & 23.9\% & 29.6\% & 25.9\% & - & - & \usym{2718} & \usym{2718} & \usym{2718} & \usym{2718} & \usym{2718} & \usym{2718} \\
    PETR \cite{liu2022petr} & 30.2\% & 30.1\% & 27.8\% & - & - & \usym{2718} & \usym{2718} & \usym{2718} & \usym{2718} & \usym{2718} & \usym{2718} \\
    Cube RCNN \cite{brazil2023omni3d} & 36.0\% & 32.7\% & 31.9\% & 36.2\% & 15.0\% & 24.9\% & 9.5\% & 27.9\% & 12.1\% & 8.5\% & 23.3\% \\
    \midrule
    UniMODE & 40.2\% & 40.0\% & 39.1\% & 36.1\% & 22.3\% & 28.3\% & 7.4\% & 29.7\% & 12.7\% & 8.1\% & 25.5\% \\
    UniMODE* & \textbf{41.3\%} & \textbf{43.6\%} & \textbf{41.0\%} & \textbf{39.8\%} & \textbf{26.9\%} & \textbf{30.2\%} & \textbf{10.6\%} & \textbf{31.1\%} & \textbf{14.9\%} & \textbf{8.7}\% & \textbf{28.2\%} \\
    \bottomrule
    \end{tabular}}
\end{table*}

\begin{table}[tbp] 
\setlength{\tabcolsep}{3pt}
    \centering
    \caption{Detailed performance of UniMODE on sub-datasets of Omni3D.} \label{Table: Detailed Performance}
    \vspace{-0.2cm}
    \resizebox{1.0\linewidth}{!}{
    \begin{tabular}{ccccccc}
    \toprule
    Backbone & $\rm AP_{3D}^{sun} \uparrow$ & $\rm AP_{3D}^{hyp} \uparrow$  & $\rm AP_{3D}^{ark} \uparrow$ & $\rm AP_{3D}^{obj} \uparrow$ & $\rm AP_{3D}^{kit} \uparrow$ & $\rm AP_{3D}^{nus} \uparrow$ \\
    \midrule
    DLA34 & 21.0\% & 6.7\% & 42.3\% & 52.5\% & 27.8\% & 31.7\% \\
    ConvNext & \textbf{23.0\%} & \textbf{8.1\%} & \textbf{48.0\%} & \textbf{66.1\%} & \textbf{29.2\%} & \textbf{36.0\%} \\
    \bottomrule
    \end{tabular}}
\end{table}

\noindent \textbf{Dataset.} The experiments in this section are primarily performed on Omni3D \cite{brazil2023omni3d} (a large-scale monocular 3D object detection benchmark encompassing both indoor and outdoor scenes) and MM-Omni3D (our developed multi-modal 3D object detection benchmark introduced in Section~\ref{SubSec: MM-Omni3D Benchmark}). Both these two benchmarks are built upon six well-known datasets including KITTI \cite{geiger2012we}, SUN-RGBD \cite{song2015sun}, ARKitScenes \cite{baruch2021arkitscenes}, Objectron \cite{ahmadyan2021objectron}, nuScenes \cite{caesar2020nuscenes}, and Hypersim \cite{roberts2021hypersim}. Among these datasets, KITTI and nuScenes focus on urban driving scenes, which are real-world outdoor scenarios. SUN-RGBD, ARKitScenes, and Objectron primarily pertain to real-world indoor environments. Compared with outdoor datasets, the required perception ranges of indoor datasets are smaller, and the object categories are more diverse. Hypersim, distinct from the aforementioned five datasets, is a synthesized dataset. Thus, Hypersim allows for the annotation of object classes that are challenging to label in real scenes, such as transparent objects (\eg, windows) and thin objects (\eg, carpets). The evaluation metric is $\rm AP_{3D}$, which reflects the 3D Intersection over Union (IoU) between 3D box predictions and labels.

\noindent \textbf{Experimental settings.} As Omni3D is a large-scale dataset, training models on it necessitates many GPUs. For example, the authors of Cube RCNN run each experiment with 48 V100s for 4$\sim$5 days. In this work, the experiments in Section~\ref{SubSec: Monocular Performance Comparison} are performed in the high computing resource setting (the input image resolution is $1280 \times 1024$, the backbone is ConvNext-Base \cite{liu2022convnet}, all training data). Since our computing resources are limited, unless explicitly stated otherwise, the remaining experiments are conducted in the low computing resource setting (the input resolution is $640 \times 512$, the backbone is DLA34 \cite{yu2018deep} if without a special statement, fixed 20\% of all training data sampled from all 6 sub-datasets).

\begin{table*}[tbp] 
    \centering
    \caption{Performance comparison between MM-UniMODE and other multi-modal 3D object detectors.} \label{Table: Multi-modal Performance Comparison}
    \vspace{-0.2cm}
    \tabcolsep=0.1cm
    \resizebox{0.9\linewidth}{!}{
    \begin{tabular}{cc|ccccccccc }
    \toprule
    Detector & Backbone & ${\rm AP_{3D}^{sun}}\uparrow$ & ${\rm AP_{3D}^{hyp}}\uparrow$ & ${\rm AP_{3D}^{ark}}\uparrow$ & ${\rm AP_{3D}^{obj}}\uparrow$ & ${\rm AP_{3D}^{kit}}\uparrow$ & ${\rm AP_{3D}^{nus}}\uparrow$ & ${\rm AP_{3D}^{in}}\uparrow$ & ${\rm AP_{3D}^{out}}\uparrow$ & ${\rm AP_{3D}}\uparrow$ \\
    \midrule
    CMT \cite{yan2023cross} & Convext-Base & 12.3\% & 5.3\% & 36.9\% & 54.9\% & 15.7\% & 18.4\% & 19.7\% & 22.7\% & 22.8\% \\
    CMT \cite{yan2023cross} & VoVNet & 8.0\% & 4.1\% & 26.9\% & 50.5\% & 13.0\% & 17.2\% & 17.4\% & 18.9\% & 18.5\% \\
    BEVFusion \cite{liang2022bevfusion} & Convext-Base & 12.3\% & 5.6\% & 35.8\% & 59.1\% & 22.3\% & 26.1\% & 21.6\% & 26.4\% & 24.1\% \\
    BEVFusion \cite{liang2022bevfusion} & VoVNet & 11.9\% & 5.4\% & 35.7\% & 52.1\% & 23.9\% & 35.7\% & 21.8\% & 33.0\% & 25.1\% \\
    FusionFormer \cite{hu2023fusionformer} & Convext-Base & 13.8\% & 5.9\% & 37.2\% & 53.1\% & 21.3\% & 24.9\% & 23.5\% & 25.9\% & 24.7\% \\
    FusionFormer \cite{hu2023fusionformer} & VoVNet & 13.4\% & 5.7\% & 37.3\% & 53.9\% & 25.0\% & 36.6\% & 24.0\% & 35.2\% & 27.0\% \\
    \midrule
    UniMODE & Convext-Base & 15.7\% & 4.8\% & 29.2\% & 49.0\% & 13.8\% & 17.6\% & 17.3\% & 20.2\% & 19.4\% \\
    MM-UniMODE & Convext-Base & \textbf{17.1\%} & 6.1\% & 39.2\% & 57.1\% & 23.1\% & 26.2\% & 24.6\% & 27.2\% & 26.1\% \\
    MM-UniMODE & VoVNet & 16.6\% & \textbf{6.3\%}  & \textbf{39.5\%} & \textbf{59.8\%} & \textbf{27.1\%} & \textbf{39.1\%} & \textbf{25.2\%} & \textbf{36.5\%} & \textbf{28.6\%} \\
    \bottomrule
    \end{tabular}}
\end{table*}

\subsection{Monocular Performance Comparison}
\label{SubSec: Monocular Performance Comparison}

In this part, we compare the performance of the proposed detector with previous monocular 3D object detectors. Among them, Cube RCNN is the sole detector that also explores unified monocular 3D object detection and it adopts a network architecture similar to FCOS3D \cite{wang2021fcos3d}. BEVFormer \cite{li2022bevformer} and PETR \cite{liu2022petr} are two popular BEV detectors, and we reimplement them in the Omni3D benchmark to get the detection scores. Among them, BEVFormer is implemented using deformable attention \cite{zhu2020deformable} while PETR is based on the original Transformer attention \cite{carion2020end}. The performance of the other compared detectors in this experiment is obtained from Omni3D \cite{brazil2023omni3d}. All the results are given in Table~\ref{Table: Monocular Performance Comparison}. In the $2_{\rm nd} \sim 4_{\rm th}$ columns of Table~\ref{Table: Monocular Performance Comparison}, the detectors are trained using KITTI and nuScenes. These three columns reflect the detection precision on KITTI test set, nuScenes test set, and overall outdoor detection performance, respectively. The $5_{\rm th} \sim 6_{\rm th}$ columns correspond to indoor detection results. Among them, the $5_{\rm th}$ column is the performance where detectors are trained and validated on SUN-RGBD. In the $6_{\rm th}$ column, detectors are trained and evaluated by combining SUN-RGBD, ARKitScenes, and Hypersim. The $7_{\rm th} \sim 12_{\rm th}$ columns represent the overall detection performance where detectors are trained and validated utilizing all data in Omni3D. UniMODE and UniMODE* denote the proposed detectors, taking DLA34 and ConvNext-Base as the backbones, respectively. The best results given various metrics are marked in \textbf{bold}. The ``\protect\usym{2718}'' means that the model does not converge well, and the obtained performance is quite poor. The ``-'' means this result is not reported in previous literature. In addition, we present the detailed detection scores of UniMODE on various sub-datasets in Omni3D as Table~\ref{Table: Detailed Performance}.

According to the results, we can observe that UniMODE achieves the best results in all metrics. It surpasses the SOTA Cube RCNN by 4.9\% given the primary metric $\rm AP_{3D}$. Besides DLA34, we also try another backbone ConvNext-Base. This is because previous papers suggest that DLA34 is commonly used in camera-view detectors like Cube RCNN but unsuitable for BEV detectors \cite{li2022diversity}. Since UniMODE is a BEV detector, testing its performance with only DLA34 is unfair. Thus, we also test UniMODE with ConvNext-Base, and the result suggests that the performance is boosted significantly. According to the results in the last two rows in Table~\ref{Table: Detailed Performance}, the detector with ConvNext-base outperforms the detector with DLA34 by 2.7\% $\rm AP_{3D}$. Additionally, the speed of UniMODE is also promising. Test on 1 A100 GPU, the inference speeds of UniMODE under the high and low computing resource settings are 21.41 FPS and 43.48 FPS separately. Moreover, it can be observed from Table~\ref{Table: Monocular Performance Comparison} that BEVFormer and PETR do not converge well in unified detection while behaving promisingly trained with outdoor datasets. This phenomenon implies the difficulty of unifying indoor and outdoor 3D object detection. Through analysis, we find that BEVFormer obtains poor results when using data of all domains because its convergence is quite unstable, and the loss curve often boosts to a high value during training. PETR does not behave well since it implicitly learns the correspondence relation between 2D pixels and 3D voxels. When the camera parameters keep similar across all samples in a dataset like nuScenes \cite{caesar2020nuscenes}, PETR converges smoothly. Nevertheless, when trained in a dataset with dramatically changing camera parameters like Omni3D, PETR becomes much more difficult to train.

\subsection{Multi-modal Performance Comparison}
\label{SubSec: Multi-modal Performance Comparison}

In this experiment, we compare the performance MM-UniMODE with previous multi-modal counterparts on the MM-Omni3D benchmark. Since there do not exist detectors that are designed for the unified multi-modal 3D object detection setting in MM-Omni3D, we select three representative methods (including CMT \cite{yan2023cross}, BEVFusion \cite{liang2022bevfusion}, and FusionFormer \cite{hu2023fusionformer}) and reproduce them in MM-Omni3D. Among them, CMT adopts a detection architecture similar to the original DETR, while BEVFusion and FusionFormer are convolution-based BEV methods. All the detectors are validated with two backbones, ConvNext-Base and VoVNet \cite{lee2019energy}. The results are presented in Table~\ref{Table: Multi-modal Performance Comparison}. The result of UniMODE is also presented for the convenience of comparison. As mentioned before, the experiments in Table~\ref{Table: Multi-modal Performance Comparison} are performed in the low computing resource setting. The best result of every metric is marked in \textbf{bold}.  We can find that MM-UniMODE obtains the best performance in all metrics. The results of MM-UniMODE are significantly better than UniMODE, which confirms that incorporating depth sensor information as input benefits detection accuracy.  Additionally, although compared with ConvNext-Base, the VoVNet backbone improves the outdoor results of BEVFusion, FusionFormer, and MM-UniMODE significantly, VoVNet degrades the precision of CMT, which is mainly because of the model structure difference. Moreover, comparing the results in Table~\ref{Table: Monocular Performance Comparison} and Table~\ref{Table: Multi-modal Performance Comparison}, although CMT is a multi-modal detector extended from the monocular detector PETR, the training of PETR collapses while CMT achieves much better results, which implies that the depth information can stabilize the training of unified 3D object detectors. We speculate this is because the depth information addresses the ill-posed nature of monocular 3D object detection.

\begin{table}[t] 
\setlength{\tabcolsep}{3pt}
    \centering
    \caption{Ablation studies on the proposed strategies in UniMODE.} \label{Table: Ablation Study}
    \vspace{-0.2cm}
    \resizebox{1.0\linewidth}{!}{
    \begin{tabular}{cccccccc}
    \toprule
    PH & UBG & SBFP & UDA & $\rm AP_{3D}^{in} \uparrow$ & $\rm AP_{3D}^{out} \uparrow$ & $\rm AP_{3D} \uparrow$ & \textbf{Improvement} \\
    \midrule
    & & & & 10.9\% & 14.3\% & 12.3\% & - \\
    \checkmark & & & & 13.4\% & 22.2\% & 15.9\% & 3.6\%$\uparrow$ \\
    \checkmark & \checkmark & & & 14.0\% & 23.8\% & 16.6\% & 0.7\%$\uparrow$ \\
    \checkmark & \checkmark & \checkmark & & 13.4\% & 23.7\%  & 16.6\%  & 0.0$\%\uparrow$ \\
    \checkmark & \checkmark & \checkmark & \checkmark & 14.8\% & 24.5\% & 17.4\% & 0.8\%$\uparrow$ \\
    \bottomrule
    \end{tabular}}
\end{table}

\subsection{Ablation Studies}
\label{SubSec: Ablation Study}

\noindent \textbf{UniMODE key designs.}
In this experiment, we ablate the effectiveness of the proposed strategies in UniMODE, including proposal head (PH), uneven BEV grid (UBG), sparse BEV feature projection (SBFP), and unified domain alignment (UDA). The experimental results are reported in Table~\ref{Table: Ablation Study}.  Specifically, the last column of Table~\ref{Table: Ablation Study} presents the improvement of each row compared with the top row. $\rm AP_{3D}^{in}$ and $\rm AP_{3D}^{out}$ reflect the indoor and outdoor detection performance, respectively. Notably, although SBFP does not boost detection precision, it reduces the computational cost of the BEV feature projection by 82.6\%. Overall, we can observe that all these strategies are very effective. Among them, the proposal head boosts the result by the most significant margin. Specifically, the proposal head enhances the overall detection performance metric $\rm AP_{3D}$ by 3.6\%. Meanwhile, the indoor and outdoor detection metrics $\rm AP_{3D}^{in}$ and $\rm AP_{3D}^{out}$ are boosted by 2.5\% and 7.9\% separately. As discussed in Section~\ref{SubSec: Two-Stage Detection Architecture}, the proposal head is quite effective because it stabilizes the convergence process of UniMODE and thus favors detection accuracy. The collapse does not happen after using the proposal head. In addition, UBG and UDA are two plug-and-play strategies, which are not only effective in improving performance but also easy to implement. Furthermore, although the sparse BEV feature projection strategy does not improve the detection precision, it reduces the projection cost by 82.6\%.

\begin{table}[t] 
    \centering
    \caption{Ablation studies on grid size and depth bin split strategy.} \label{Table: Study on BEV Feature}
    \vspace{-0.2cm}
    \resizebox{0.85\linewidth}{!}{
    \begin{tabular}{ccccc}
    \toprule
    Grid Size (m) & Depth Bin & $\rm AP_{3D}^{in} \uparrow$ & $\rm AP_{3D}^{out} \uparrow$ & $\rm AP_{3D} \uparrow$ \\
    \midrule
    1 & Even & 14.8\% & 24.5\% & 17.4\% \\
    1 & Uneven & 12.1\% & 22.6\% & 15.3\% \\
    0.5 & Even & 15.4\% & 25.5\% & 18.1\% \\
    2 & Even & 14.0\% & 23.9\% & 16.5\% \\
    \bottomrule
    \end{tabular}}
\end{table}

\begin{table}[t] 
    \centering
    \caption{Ablation study on $\tau$ in sparse BEV feature projection.} \label{Table: study on sparse feature projection}
    \vspace{-0.2cm}
    \resizebox{0.8\linewidth}{!}{
    \begin{tabular}{ccccc}
    \toprule
    $\tau$ & Remove Ratio (\%) & $\rm AP_{3D}^{in} \uparrow$ & $\rm AP_{3D}^{out} \uparrow$ & $\rm AP_{3D} \uparrow$ \\
    \midrule
    0 & 0.0 & 14.9\% & 24.7\% & 17.4\% \\
    1e-3 & 82.6 & 14.8\% & 24.5\% & 17.4\% \\
    1e-2 & 94.3 & 12.1\% & 21.9\% & 15.0\% \\
    1e-1 & 98.3 & 4.7\% & 3.6\% & 4.7\% \\
    \bottomrule
    \end{tabular}}
\end{table}

\noindent \textbf{Analysis of uneven BEV grid.}
We study the effect of BEV feature grid size and depth bin split strategy in uneven BEV grid design, and the results are presented in Table~\ref{Table: Study on BEV Feature}. When the depth bin split is uneven, we split the depth bin range following Eq.~\ref{Eq1}. Comparing the $1_{\rm st}$ and $2_{\rm nd}$ rows of results in Table~\ref{Table: Study on BEV Feature}, we can find that uneven depth bin deteriorates detection performance. We speculate this is because this strategy projects more points in closer BEV grids while fewer points in farther grids, which further increases the imbalanced distribution of projection features. Additionally, comparing the $1_{\rm st}$, $3_{\rm rd}$, and $4_{\rm th}$ rows of results in Table~\ref{Table: Study on BEV Feature}, it is observed that smaller BEV grids lead to better performance. Due to the consideration of computing resource budget, We set the BEV grid size to 1m rather than 0.5m in all the other experiments. According to the results of this experiment, we can infer that if we further decrease the size of BEV grids, the performance of UniMODE can be continuously boosted compared with the current performance. 

\noindent \textbf{Analysis of sparse BEV feature projection.} As mentioned in Section~\ref{SubSec: Sparse BEV Feature Projection}, the BEV feature projection process is computationally expensive. To reduce this cost, we propose to remove unimportant projection points. Although this strategy enhances network efficiency significantly, it could deteriorate detection accuracy and convergence stability due to the reduction of retained feature. Hence, this is exactly a trade-off between efficiency and detection precision. In this part, we study this trade-off through experiments. Specifically, as introduced in Section~\ref{SubSec: Sparse BEV Feature Projection}, we remove unimportant projection points based on a pre-defined hyper-parameter $\tau$. Therefore, in this experiment, we adjust the value of $\tau$ to analyze how the removed projection point ratio affects performance, and the results are reported in Table~\ref{Table: study on sparse feature projection}.

It can be observed from Table~\ref{Table: study on sparse feature projection} that when $\tau$ is 0, which means no feature is discarded, the best performance across all rows is arrived. When we set $\tau$ to $1e^{-3}$, about 82.6\% of the feature is discarded while the performance of the detector remains very similar to the one with $\tau=0$. This phenomenon suggests that the discarded feature in the setting $\tau=1e^{-3}$ is unimportant for maintaining the final detection accuracy. Then, when we increase $\tau$ to $1e^{-2}$ and $1e^{-1}$, we can find that the corresponding detection performances drop dramatically. This observation indicates that when we discard superfluous features, the detection precision and even training stability would be influenced significantly. Combining all the observations, we set $\tau$ to $1e^{-3}$ and drop $82.6\%$ of unimportant features in UniMODE, which reduces the computational cost by $82.6\%$ while maintaining similar performance as the one without dropping features. 

\begin{table}[t] 
\setlength{\tabcolsep}{2.5pt}
    \centering
    \caption{Analysis on the effectiveness of DALN.} \label{Table: DALN}
    \vspace{-0.2cm}
    \resizebox{0.85\linewidth}{!}{
    \begin{tabular}{c|cc|cc|cc}
    \toprule
    \multirow{2}{*}{Method} & \multicolumn{2}{c|}{None} & \multicolumn{2}{c|}{DR} & \multicolumn{2}{c}{DALN} \\
    & ${\rm AP^{ark}_{3D}}\uparrow$ & ${\rm AP^{sun}_{3D}}\uparrow$ & ${\rm AP^{ark}_{3D}}\uparrow$ & ${\rm AP^{sun}_{3D}}\uparrow$ & ${\rm AP^{ark}_{3D}}\uparrow$ & ${\rm AP^{sun}_{3D}}\uparrow$ \\
    \midrule
    Result & 33.6\% & 12.3\% & 33.9\% & 12.1\% & 35.0\% & 13.0\% \\
    \bottomrule
    \end{tabular}}
\end{table}

\begin{table}[t] 
\setlength{\tabcolsep}{1pt}
    \centering
    \caption{Ablation study on the MIC strategy.} \label{Table: ablation study on MIC}
    \vspace{-0.2cm}
    \tabcolsep=0.1cm
    \resizebox{1.0\linewidth}{!}{
    \begin{tabular}{cc|ccccccc}
    \toprule
    $L_{P \rightarrow I}$ & $L_{I \rightarrow P}$ & ${\rm AP_{3D}^{sun}}$ & ${\rm AP_{3D}^{hyp}}$ & ${\rm AP_{3D}^{ark}}$ & ${\rm AP_{3D}^{obj}}$ & ${\rm AP_{3D}^{kit}}$ & ${\rm AP_{3D}^{nus}}$ & ${\rm AP_{3D}}$ \\
    \midrule
    & & 16.5\% & 5.7\% & 36.7\% & 50.8\% & 20.5\% & 23.8\% & 23.8\% \\
    \checkmark & & 16.3\% & 5.9\% & 37.8\% & 56.0\% & 22.7\% & 27.4\% & 25.1\% \\
    & \checkmark & 17.1\% & 5.9\% & 38.3\% & 56.4\% & 22.7\% & 26.3\% & 25.7\% \\
    \checkmark & \checkmark & 17.1\% & 6.1\% & 39.2\% & 57.1\% & 23.1\% & 26.2\% & 26.1\% \\
    \bottomrule
    \end{tabular}}
\end{table}

\noindent \textbf{Analysis of DALN.} In this experiment, we validate the effectiveness of DALN through comparing the performances of the naive baseline without any domain adaptive strategy, the baseline predicting dynamic parameters with direct regression (DR) \cite{park2019semantic}, and the baseline with DALN (our proposed). All these models are trained using only the ARKitScenes dataset. Afterwards, they are evaluated on ARKitScenes (in-domain evaluation) and SUN-RGBD (out-of-domain evaluation) separately. The experimental result is presented in Table~\ref{Table: DALN}. It can be observed that DR could degrade the detection accuracy. This is because directly predicting accurate dynamic parameters is challenging and harms the training stability. By contrast, DALN boosts the performance significantly, which reveals the zero-shot out-of-domain effectiveness of DALN.

\noindent \textbf{Analysis of MIC.}
In this part, we verify the effectiveness of MIC used for image and point feature fusion in MM-UniMODE. According to the description in Section~\ref{SubSec: Mutual Information Collaboration}, there are two kinds of guidance losses in MIC, \ie, $L_{P \rightarrow I}$ and $L_{I \rightarrow P}$. Among them, $L_{P \rightarrow I}$ utilizes precise depth information to guide the extraction of ambiguous image feature, and $L_{I \rightarrow P}$ employs informative image representation to aid the feature lack in grids that do not contain depth sensor points. In this experiment, we evaluate the effectiveness of  $L_{P \rightarrow I}$ and $L_{I \rightarrow P}$ individually. The results are reported in Table~\ref{Table: ablation study on MIC}. As presented, both $L_{P \rightarrow I}$ and $L_{I \rightarrow P}$ boost the baseline detector performance significantly. Among them, $L_{I \rightarrow P}$ plays a more important role, which boosts the ${\rm AP_{3D}}$ metric by 1.9\%. Moreover, $L_{P \rightarrow I}$ enhances ${\rm AP_{3D}}$ by 1.3\%, which is also a significant improvement. When we combine $L_{I \rightarrow P}$ and $L_{P \rightarrow I}$, the overall detection precision is improved by 2.3\%. The results well confirm the effectiveness of the proposed MIC module and guidance loss items.

\begin{table}[t]
\setlength{\tabcolsep}{2.5pt}
    \centering
    \caption{Cross-domain evaluation in indoor sub-datasets.} \label{Table: Cross-domain Evaluation}
    \vspace{-0.2cm}
    \resizebox{0.99\linewidth}{!}{
    \begin{tabular}{c|ccc|ccc}
    \toprule
    \multirow{2}{*}{Train} & \multicolumn{3}{c|}{Zero-Shot} & \multicolumn{3}{c}{$\delta$-Tune} \\
    & $\rm AP_{3D}^{hyp} \uparrow$ & $\rm AP_{3D}^{sun} \uparrow$ & $\rm AP_{3D}^{ark} \uparrow$ & $\rm AP_{3D}^{hyp} \uparrow$ & $\rm AP_{3D}^{sun} \uparrow$ & $\rm AP_{3D}^{ark} \uparrow$  \\
    \midrule
    Hypersim & 14.7\% & 5.6\% & 3.6\% & 14.7\% & 18.5\%  & 18.9\%  \\
    SUN-RGBD & 3.0\% & 28.5\% & 8.8\% & 7.5\% & 28.5\% & 27.2\% \\
    ARKitScenes & 4.2\% & 13.0\% & 35.0\% & 10.4\% & 22.8\% & 35.0\% \\
    \bottomrule
    \end{tabular}}
\end{table}

\begin{table}[t]
\setlength{\tabcolsep}{2.5pt}
    \centering
    \caption{Analysis on how training data volume affects out-of-domain generalization performance.} \label{Table: out-of-domain benefit}
    \vspace{-0.2cm}
    \resizebox{0.75\linewidth}{!}{
    \begin{tabular}{cc}
    \toprule
    Training Data & ${\rm AP_{3D}^{sun}} \uparrow$ \\
    \midrule
    SUN-RGBD & 17.3\% \\
    SUN-RGBD \& Objectron \& ARKitScenes & 21.5\% \\
    Objectron \& ARKitScenes & 19.6\% \\
    Objectron \& ARKitScenes \& KITTI & 20.1\% \\
    \bottomrule
    \end{tabular}}
\end{table}

\begin{table}[t]
\setlength{\tabcolsep}{2.5pt}
    \centering
    \caption{Analysis on how MM-Omni3D pre-training affects the performance in the KITTI test set.} \label{Table: pre-train study on kitti}
    \vspace{-0.2cm}
    \resizebox{0.7\linewidth}{!}{
    \begin{tabular}{cccc}
    \toprule
    Training & ${\rm AP_{3D}^{easy}} \uparrow$ & ${\rm AP_{3D}^{mode.}} \uparrow$ & ${\rm AP_{3D}^{hard}} \uparrow$ \\
    \midrule
    KITTI & 90.0\% & 81.2\% & 75.4\%  \\
    DD3D-KITTI & 89.8\% & 82.6\% & 79.4\% \\
    MM-Omni3D & 91.2\% & 84.8\% & 80.7\% \\
    MM-Omni3D* & 91.9\% & 85.2\% & 81.2\% \\
    \bottomrule
    \end{tabular}}
\end{table}

\subsection{Monocular Cross-domain Evaluation}
\label{SubSec: Cross-domain Evaluation}

We evaluate the generalization ability of UniMODE in this part by conducting cross-domain evaluation between various indoor sub-datasets in Omni3D. Specifically, we train a detector on an indoor sub-dataset of Omni3D and test the performance of this detector on different other sub-datasets. The experiments are conducted in two settings, \ie, the zero-shot setting and $\delta$-tune setting. In the zero-shot setting, the test domain is completely unseen. In the $\delta$-tune setting, 1\% of the training set data from the test domain is used to fine-tune the Query FFN in UniMODE for 1 epoch. The experimental results are presented in Table~\ref{Table: Cross-domain Evaluation}.

According to the results in the $2_{\rm st} \sim 4_{\rm th}$ columns of Table~\ref{Table: Cross-domain Evaluation}, we can find that when a detector is trained and validated in the same indoor sub-dataset, its performance is promising. However, when evaluated on another completely unseen sub-dataset, the accuracy is limited, especially for the virtual dataset Hypersim. This is partly because monocular 3D depth estimation is an ill-posed problem. When the training and validation data belong to differing domains, predicting depth accurately is challenging, especially for the virtual dataset Hypersim. Then, we introduce another testing setting, $\delta$-tuning. In this setting, if a detector is tested on another domain different from the training domain, the Query FFN is fine-tuned by 1\% of training data from the test domain. 
The results of the $\delta$-tuning setting are reported in the $5_{\rm st} \sim 7_{\rm th}$ columns of Table~\ref{Table: Cross-domain Evaluation}. We can observe that when fine-tuned with only a handful of data, the performance of UniMODE becomes much more promising. This result suggests that the superiority of UniMODE in serving as a foundational model. It can benefit practical applications by only incorporating a little training data from the test domain.

\subsection{More Analysis on MM-Omni3D}
\label{SubSec: Analysis on MM-Omni3D}

\noindent \textbf{Out-of-domain generalization.} In this part, we analyze how the training data volume affects the out-of-domain detection accuracy of MM-UniMODE. Specifically, we train MM-UniMODE using only the SUN-RGBD subset in MM-Omni3D, all the indoor subsets in MM-Omni3D, all indoor subsets except SUN-RGBD, and all indoor subsets besides KITTI, respectively. Then, we verify the performances of these models with the SUN-RGBD test set, and the results are reported in Table~\ref{Table: out-of-domain benefit}. It can be observed that the detector trained based on all the indoor subsets except SUN-RGBD outperforms the detector trained solely with SUN-RGBD, which indicates the promising out-of-domain generalization capability of MM-UniMODE. This observation also suggests that training detectors with more abundant data from more diverse domains generally improves the out-of-domain generalization abilities of the trained detectors.

\noindent \textbf{MM-Omni3D as Pre-training dataset.} In this part, we first delve into how pre-training using MM-Omni3D affects the performance on the KITTI test set, and the results are presented in Table~\ref{Table: pre-train study on kitti}. The 4 rows of results in this table correspond to the models trained with KITTI, first pre-trained on DD3D \cite{park2021pseudo} and then fine-tuned using KITTI, solely trained with MM-Omni3D, first pre-trained on MM-Omni3D and then fine-tuned using KITTI, respectively. We can observe that MM-Omni3D serves as a more promising pre-training dataset than DD3D, which is a large-scale depth estimation dataset and has been commonly used by previous works. Therefore, besides serving as a multi-modal 3D object detection benchmark, MM-Omni3D also shows promise as a pre-training dataset and has the potential to favor different kinds of 3D perception tasks.

\begin{figure*}[tb]
    \centering
    \includegraphics[width=1.0\textwidth]{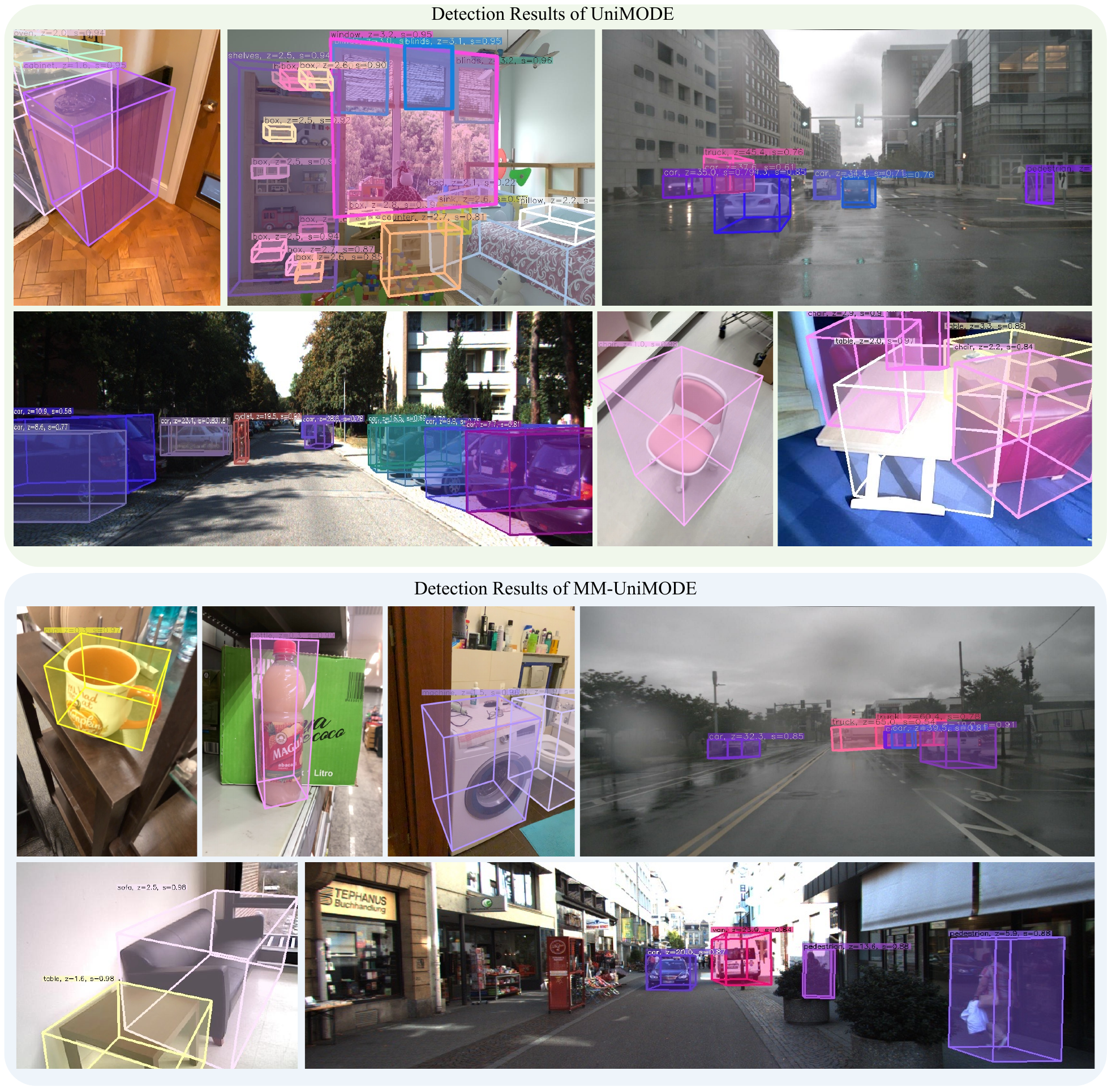}
    \caption{Visualization of the UniMODE and MM-UniMODE detection results in various scenarios. It can be observed that UniMODE achieves promising detection precision and incorporating depth information further boosts the performance.}\label{Fig: Vis}
\end{figure*}

\subsection{Visualization}
\label{SubSec: Visualization}
In Fig.~\ref{Fig: Vis}, we showcase some detection results of UniMODE and MM-UniMODE in various scenarios. It can be observed that both these two detectors can accurately capture 3D object boxes under both complex indoor and outdoor scenarios. UniMODE realizes promising 3D object detection primarily through algorithm design. Then, via integrating the improvement from the data perspective, the detection results of MM-UniMODE are generally more precise due to the adoption of depth sensor information as input. 

\section{Conclusion}
\label{Sec: Conclusion}

In this work, we have delved into the unified 3D object detection problem in both the monocular and multi-modal settings. In terms of unified monocular 3D object detection, we have proposed a two-stage detection architecture, an uneven BEV grid split strategy, and a domain alignment method to handle the unstable training problem. In addition, a sparse BEV feature projection technique has been developed to alleviate the computing burden. Combining these techniques, a detector named UniMODE has been obtained and it has achieved SOTA performance in the Omni3D benchmark. To explore the unified multi-modal 3D object detection problem, we have built the first corresponding benchmark named MM-Omni3D. Afterwards, we have extended UniMODE to the multi-modal detector MM-UniMODE based on the the proposed module MIC. Extensive experiments have been performed to verify the effectiveness of the various proposed techniques and demonstrated the outstanding performances of UniMODE and MM-UniMODE. In addition, the experimental results also have confirmed MM-Omni3D can serve as a promising pre-training or co-training dataset to improve the generalization capabilities of 3D object detectors.

%\vfill

{\small
\bibliographystyle{ieeetr}
\bibliography{reference}
}

\begin{IEEEbiography}[{\includegraphics[width=1in,height=1.25in,clip,keepaspectratio]{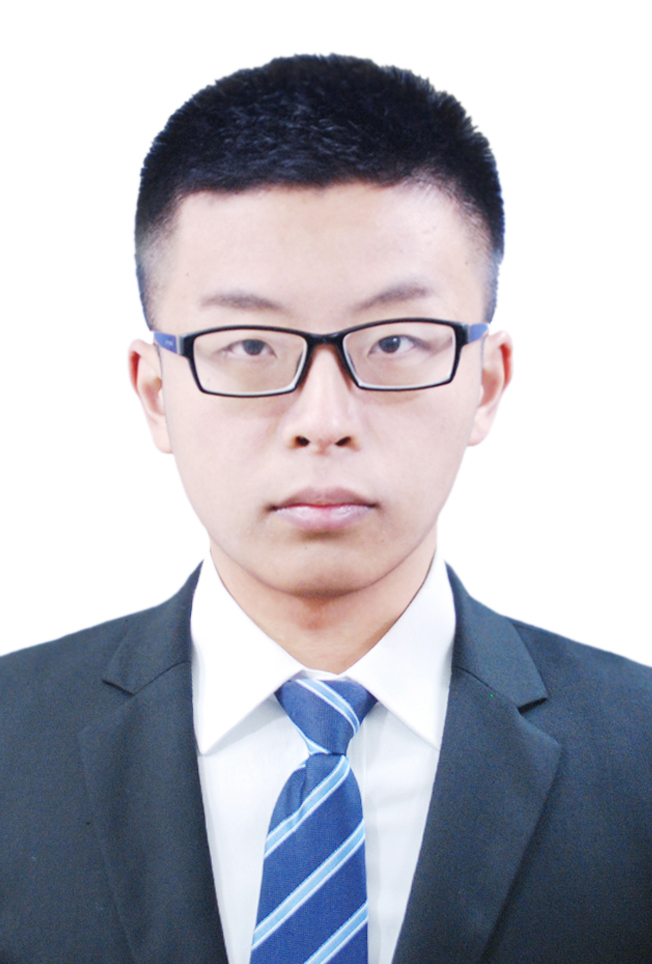}}]{Zhuoling Li}
received the B.S. degree from Huazhong University of Science and Technology in 2020 and received the M.S. degree from Tsinghua University in 2023. He is currently pursuing the Ph.D. degree in the University of Hong Kong. He serves as a reviewer for CVPR, ICCV, ECCV, ICML, NeurIPS, etc. His current research interests include embodied intelligence and 3D vision perception.
\end{IEEEbiography}

\begin{IEEEbiography}[{\includegraphics[width=1in,height=1.25in,clip,keepaspectratio]{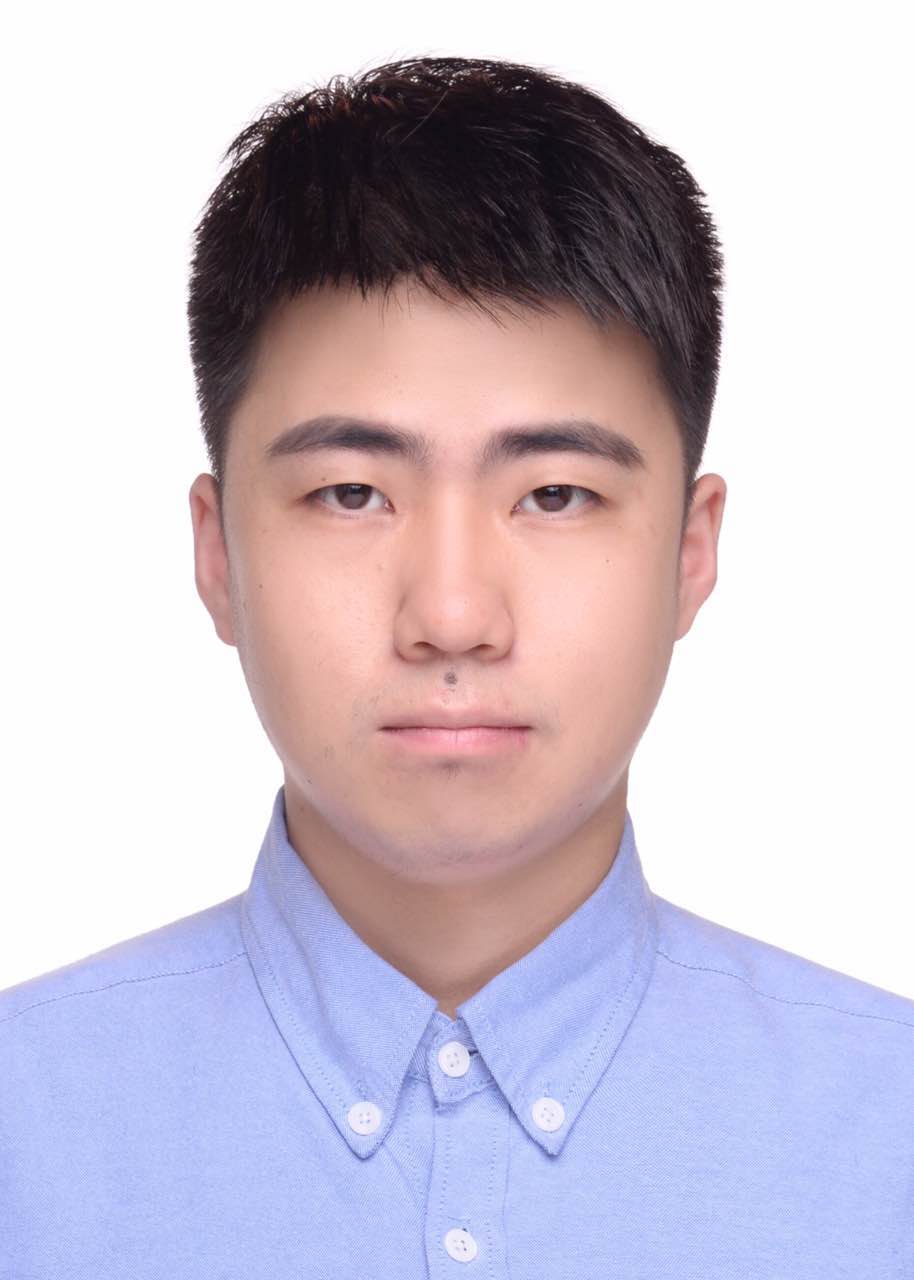}}]{Xiaogang Xu}
is currently a ZJU100 Professor in Zhejiang University. Before that, he was a research fellow in the Chinese University of Hong Kong. He obtained his Ph.D. degree from the Chinese University of Hong Kong in 2022. He received his bachelor degree from Zhejiang University. He obtained the Hong Kong PhD Fellowship in 2018. He serves as a reviewer for CVPR, ICCV, ECCV, AAAI, ICLR, NIPS, IJCV. His research interest includes deep learning, generative adversarial networks, adversarial attack and defense, etc.
\end{IEEEbiography}

\begin{IEEEbiography}[{\includegraphics[width=1in,height=1.25in,clip,keepaspectratio]{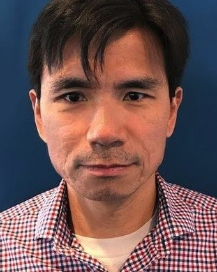}}]{Ser-Nam Lim} concentrates on Computer Vision and the field of AI. He pursued and earned a PhD at the University of Maryland, College Park, in 2005. Ser-Nam spent a decade at GE Research working on different areas of Computer Vision including video recognition, 3D reconstruction, representation, and visual matching.  At GE Research, Ser-Nam started as a Research Scientist, then took a role as the Computer Vision Lab Manager, and finished his career there as a Senior Principal and Director. Ser-Nam then took a role managing multiple AI teams at Meta that conduct research in Computer Vision, NLP and other areas of AI with a focus on applying and scaling to huge amounts of data on the Meta platform. Ser-Nam’s work has accomplished impactful production outcomes in ensuring safety in the Aviation and Power industry as well as detecting misinformation and other integrity issues on the Meta platform. At the end of his career at Meta, Ser-Nam leads projects focused on AI for user to content recommendations, as well as search engines that include the intersection of Large Language Models (LLM) and Computer Vision. After Meta, Ser-Nam joined the Computer Science faculty at University of Central Florida (UCF) as a tenured Associate Professor. His group at UCF conducts research in image and video generation, AI for Augmented Reality, visual-language representation and understanding and other major topics in AI. Ser-Nam has published over 100 peer-refereed papers, with more than half in top AI venues.
\end{IEEEbiography}

\begin{IEEEbiography}[{\includegraphics[width=1in,height=1.25in,clip,keepaspectratio]{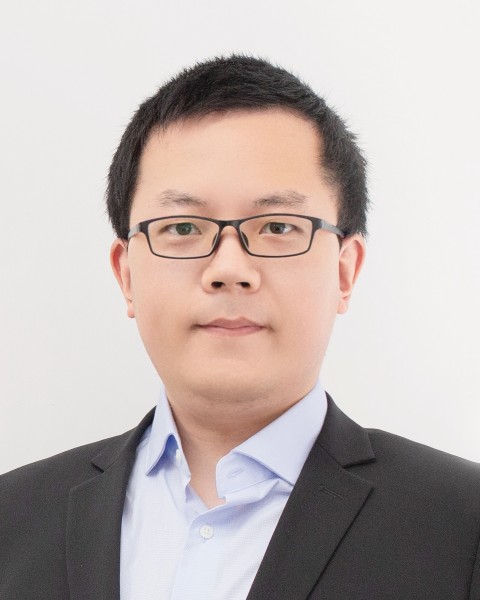}}]{Hengshuang Zhao}
is an Assistant Professor at the Department of Computer Science of The University of Hong Kong. Previously, he was a postdoctoral researcher at Massachusetts Institute of Technology and University of Oxford. He received the Ph.D. degree in Computer Science and Engineering from The Chinese University of Hong Kong. His general research interests cover the broad area of computer vision, machine learning and artificial intelligence, with special emphasis on building intelligent visual systems. He is recognized as one of the most influential scholars in computer vision by AI 2000 in 2022, 2023, and 2024. He is an Area Chair for CVPR'23, NeurIPS'23, WACV'23, CVPR'24, ECCV'24, NeurIPS'24, ACMMM'24, WACV'25, 3DV'25, ICLR'25, CVPR'25, a Senior Program Committee for AAAI'23, AAAI'24, AAAI'25, an Associate Editor for Pattern Recognition, and a Guest Editor for IEEE Transactions on Circuits and Systems for Video Technology (TCSVT). He is a member of the IEEE.
\end{IEEEbiography}

\end{document}